\documentclass[10pt,letterpaper]{article}

\def\ARXIVSKIP{}

\usepackage[letterpaper,margin=1in]{geometry}

\ifx\XeTeXrevision\undefined
  \usepackage[utf8]{inputenc}
\fi
\usepackage[T1]{fontenc}
\usepackage{newtxtext}
\usepackage{newtxmath}
\usepackage{amsmath,mathtools}

\usepackage{amssymb}

\usepackage{amsthm}
\usepackage{graphicx}
\usepackage{booktabs}
\usepackage{array}
\usepackage{tabularx}
\usepackage{rotating}
\usepackage{multirow}
\usepackage{xcolor}
\usepackage[colorlinks=true,allcolors=blue!55!black]{hyperref}
\usepackage{tcolorbox}
\usepackage{caption}
\usepackage{microtype}

\graphicspath{{./}{../figures/}{./figures/}}

\newsavebox{\fitbox}
\newcommand{\fitwidth}[1]{%
  \sbox{\fitbox}{#1}%
  \ifdim\wd\fitbox>\linewidth
    \resizebox{\linewidth}{!}{\usebox{\fitbox}}%
  \else
    \usebox{\fitbox}%
  \fi}

\newcommand{\fitmath}[1]{%
  \sbox{\fitbox}{$\displaystyle #1$}%
  \ifdim\wd\fitbox>\linewidth
    \resizebox{\linewidth}{!}{\usebox{\fitbox}}%
  \else
    \usebox{\fitbox}%
  \fi}

\newtcolorbox{keybox}{
  colback=black!3!white, colframe=black!45!white,
  boxrule=0.4pt, arc=1pt, left=5pt, right=5pt, top=4pt, bottom=4pt,
  fontupper=\small}

\theoremstyle{plain}

\theoremstyle{remark}

\title{\texttt{MA-Means}: Bounded-Memory $k$-Means via Elastic Multi-Candidate Contraction,\\[2pt]
Coordinate-Wise Over-Relaxation, and Zero-Error Out-of-Core Bypass}

\author{%
  Meng Zhu
}

\hypersetup{%
  pdftitle={{MA-Means}: Bounded-Memory k-Means via Elastic Multi-Candidate Contraction, Coordinate-Wise Over-Relaxation, and Zero-Error Out-of-Core Bypass},
  pdfauthor={Meng Zhu},
  pdfsubject={Large-scale k-means clustering with bounded memory},
  pdfkeywords={k-means, clustering, vector quantization, GPU, Triton, over-relaxation, Ward contraction, out-of-core},
}

\begin{document}

\maketitle

\begin{abstract}
Large-scale $k$-means faces a memory--convergence trilemma: full-batch Lloyd requires
$O(nKD)$ work per step and traps in local minima; memoryless mini-batch EMA solvers
(Sculley's MBKM and Meta FAIR's \texttt{e-means}) are effective
on pre-cleaned uniform embeddings within fixed-basin linear updates ($w\in[0,1]$) using
only scalar counts $N_j$, yet cannot cross uphill topology barriers, destroy rare
heavy-tail modes under mass-gated revival (proportional donor split), and provide no
out-of-core filtering; and staleness-reduction methods maintain an unbounded $\Theta(n)$
sample table reaching tens of gigabytes at web scale. Extending the EMA
scaling-law and closing-mean foundation beyond single-trajectory convex updates, we
introduce \texttt{MA-Means}, a bounded-memory clustering framework holding
$O(1)$-in-$n$ multi-horizon statistics ($0.1\text{--}51\text{ MB}$ flat for
$n=10^4\text{--}10^{10}$) inside a zero-copy Triton BF16 + fused coordinate-wise update
+ async C++17 CPU pipeline. \texttt{MA-Means} unites three mechanisms: (i) an
\textbf{adaptive elastic multi-candidate pool ($K\to K'\to K$)} that spawns candidates on
high-distortion modes, co-evolves them across non-monotone barriers, and contracts
$K'\to K$ in $O(\log(K'-K))$ GPU waves via Benz\'ecri reciprocal-nearest-neighbor
Huygens--Ward merging alongside $O(K)$ distortion-aware revival (Theorems 1 \& 1.3,
Lemma 1.2); (ii) \textbf{statistic-driven coordinate-wise Aitken over-relaxation
($w_{j,d}\in\mathbb{R}^D$, $w_{j,d}>1$)} backed by a $T$-step tracking boundary theorem
proving the $O(1/(T^2\rho_{\min}))$ optimality of fixed-basin linear EMA
schedules ($w\in[0,1]$) while proving that moving Voronoi boundaries leave bias
$\ge\lambda_{j,d}$ eliminated in one step by $w_{j,d}^*=(1-\lambda_{j,d})^{-1}$
(Theorem 4, Corollary 4.1, Proposition 4.2); and (iii) coordinate-balanced
\texttt{int8\_chord} checkpoints with monotonic jump latches for deterministic zero-error
host$\leftrightarrow$device transfer refusal (Theorems 3 \& 5). Across five public
benchmarks (SIFT1M, Deep10M up to $9.97\text{M}$, ImageNet-CLIP 768d, Cohere-10M 768d,
NYC-Taxi-20M 4D; 10 paired seeds for Layer-2 and 5 for Layer-1, zero per-dataset tuning)
and four synthetic suites up to $n=10^7$ against 14 baselines, \texttt{MA-Means} wins
12 of 15 Layer-1 cells at 5 epochs (the three 1M-row $K{=}4096$ cells are within
$+0.05\%$--$+0.07\%$ MSE), including $1.69\times\text{--}1.78\times$ equal-MSE speedup at
3 epochs vs.\ 5-epoch \texttt{e-means} on full $9.97\text{M}$ Deep10M and
$1.62\times\text{--}1.66\times$ on ImageNet-CLIP; at equal 5 epochs the full-Deep10M
$K{=}1024$ gain is $1.08\times$. In Layer-2 BF16 it is
$1.32\times\text{--}1.90\times$ faster at equal MSE on SIFT1M/Deep10M/Cohere
($K{=}1024$), wins $10/10$ seeds on ImageNet-CLIP ($-0.61\%$ MSE at $K{=}1024$, beating
25ep \texttt{e-means} and exact Hamerly in $3.64\times$ less time than 5ep
\texttt{e-means}; $-0.26\%$ at $K{=}4096$, $1.58\times$ faster), cuts NYC-Taxi-20M MSE
by $54.8\%\text{--}63.3\%$ vs.\ \texttt{e-means} ($>4.6\times\text{--}7.6\times$ faster
at equal MSE; at $K{=}4096$, $\sim150$--$300\times$ faster at $+5\%$--$10\%$ MSE (5 epochs) or $\sim5\times$ faster within $\pm6\%$ MSE (200 epochs); at $K{=}1024$, $\sim90$--$120\times$ faster at $+8\%$--$16\%$ MSE and no budget closes the gap),
runs in $0.59\times\text{--}0.77\times$ time at $n=10\text{M}$ with $-5.3\%$ to $-87.3\%$
lower MSE across all four $10\text{M}$ synthetic suites, and eliminates
$14.4\%\text{--}26.4\%$ of out-of-core host$\leftrightarrow$device traffic ($0$ false
positives, $-19.8\%$ MSE).
\end{abstract}

\section{\texorpdfstring{Introduction}{Introduction}}

Partitioning $n$ vectors $X = \{x_1, \dots, x_n\} \subset \mathbb{R}^D$ into $K$ centroids $C = \{c_1, \dots, c_K\} \subset \mathbb{R}^D$ to minimize squared Euclidean distortion $\mathcal{L}(C) = \frac{1}{n}\sum_{i=1}^n \min_{j \in [K]} \|x_i - c_j\|_2^2$ underpins vector quantization (\texttt{IVF}/\texttt{PQ} \cite{ref17,ref18}), web-scale pretraining deduplication (\texttt{SemDeDup} \cite{ref19}, \texttt{DINOv2} \cite{ref20}), discrete codebook learning (\texttt{VQ-\allowbreak{}VAE} \cite{ref21}), and heavy-tailed LLM activation/KV-cache quantization \cite{ref22,ref23}. Yet existing solvers occupy three incompatible corners of state retention, optimization geometry, and hardware utilization (\textbf{Table 1}):

\begin{table}[htbp]
\centering
\caption*{Table 1: Comparison of large-scale $k$-means paradigms by retained state, update geometry, and hardware model.}
\label{tab:1}
\small
\setlength{\tabcolsep}{4pt}
\fitwidth{%
\begin{tabular}{@{}llcll@{}}
\toprule
\textbf{Paradigm} & \textbf{Representative Methods} & \textbf{Retained State} & \textbf{Key Limitation / Capability} & \textbf{Hardware Model} \\
\midrule
\textbf{Memoryless Full-Batch} & Lloyd \cite{ref1}, \texttt{k-\allowbreak{}means++} \cite{ref2}, Flash-KMeans \cite{ref9} & $0$ (beyond $C$) & $O(nKD)$ per step; traps in local minima under random init & Homogeneous GPU GEMM \\
\textbf{1st-Order Linear EMA} & Sculley MBKM \cite{ref3}, Meta FAIR \texttt{e-\allowbreak{}means} \cite{ref8} & $O(K)$ scalars ($N_j$) & Fixed-$K$, $w \in [0,1]$; mass splits hurt heavy tails; no OOC bypass & Homogeneous GPU GEMM \\
\textbf{Unbounded Sample-Memory} & Nested MBKM \cite{ref6}, \texttt{SR-\allowbreak{}MBKM} \cite{ref7}, Elkan \cite{ref10} & $\Theta(n)$ ints/bounds & $37.3\text{ GB}$ RAM at $n=10^{10}$; OOM at $K\ge 4096$ & Homogeneous CPU/GPU \\
\textbf{Bounded Multi-Horizon (Ours)} & \textbf{\texttt{MA-\allowbreak{}Means} (Elastic + CoordOR + OOC)} & \textbf{$O(1)$ in $n$ (\texttt{0.1-51 MB})} & \textbf{Non-monotone $K\!\to\!K'\!\to\!K$ + $w_{j,d}\!>\!1$ + 0-FP transfer bypass} & \textbf{Fused Triton BF16 $\parallel$ C++17 CPU} \\
\bottomrule
\end{tabular}
}
\end{table}

Mini-batch EMA with scaling laws and \texttt{closing\_\allowbreak{}mean} \cite{ref8} matches 25-epoch Lloyd on smooth embeddings at a fraction of the cost. Extending EMA updates beyond fixed-$K$ convex updates ($w_t \in [0,1]$) and scalar counts $N_j$, we resolve three barriers beyond uniform embeddings or single-GPU HBM: (i) \textbf{rigid topology}---fixed-$K$ trajectories cannot cross uphill barriers, while mass-gated revival ($N_j < 0.1\bar{N}$, proportional donor split) destroys rare heavy-tail modes (\texttt{+64.4\%} median test MSE with revival on vs. off on \texttt{NYC-\allowbreak{}Taxi-\allowbreak{}20M}, 5 seeds, $K{=}1024$); (ii) \textbf{moving-boundary lag}---any convex step $w \in [0,1]$ under-relaxes shifting Voronoi walls, and \texttt{e-\allowbreak{}means}' marginal cost rises $+20\%$ from $1\text{M}$ to $10\text{M}$ on \texttt{Deep10M}; and (iii) \textbf{no out-of-core bypass}---rows exceeding HBM must traverse host$\leftrightarrow$device links every epoch.

\textbf{Contributions.} Building on exponential moving averages and scalar sufficient statistics, \texttt{MA-\allowbreak{}Means} makes four contributions:

\begin{enumerate}
\item 
\textbf{$O(1)$-in-$n$ Elastic \& Multi-Horizon Memory (\S3)}: Bounded cluster/coordinate tables (\texttt{0.1-51 MB} flat in $n$ across $D \in [4, 768]$, \textbf{Figure 1}) plus \texttt{80 B/\allowbreak{}tile} out-of-core fingerprints replace $\Theta(n)$ sample tables.
\item 
\textbf{Exact Guarantees Beyond Linear Fixed-Basin Schedules (\S4)}: We prove (i) cell-level Huygens--Ward descent and a $\frac{2}{\pi-2}$ unimodal ceiling (\textbf{Thm. 1, Lem. 1.2}) plus \textbf{Theorem 1.3} (Benz\'ecri RNN parallel wave Ward contraction in $O(\log(K'-K))$ waves and uphill-barrier escape via $K\to K'\to K$); (ii) the $O(1/(T^2\rho_{\min}))$ optimality of fixed-basin linear EMA schedules ($w\in[0,1]$) (\textbf{Thm. 4, Cor. 4.1}) alongside \textbf{Proposition 4.2} (moving Voronoi boundaries leave bias $\ge\lambda_{j,d}$ under $w\in[0,1]$, eliminated in one step by coordinate-wise Aitken $w_{j,d}^*=(1-\lambda_{j,d})^{-1}>1$); and (iii) deterministic zero-error \texttt{int8\_\allowbreak{}chord} bypass and a medium-agnostic transfer law (\textbf{Thms. 3 \& 5}).
\item 
\textbf{Pure Layer-1 Algorithmic Speedups \& $10\times$ Scaling (\S5--\S6)}: In \textbf{100\% operator-neutral Layer-1 FP32}, \texttt{MA-\allowbreak{}Means} wins \textbf{12 of 15 Layer-1 cells} at 5 epochs (the three 1M-row $K{=}4096$ cells are within \texttt{+0.05\%}--\texttt{+0.07\%} MSE), including \textbf{\texttt{1.69$\times$-1.78$\times$}} equal-MSE speedup at 3 epochs vs. 5-epoch \texttt{e-\allowbreak{}means} on \textbf{full $9.97\text{M}$ \texttt{Deep10M}} and \textbf{\texttt{1.62$\times$-1.66$\times$}} on \texttt{ImageNet-\allowbreak{}CLIP}. In Layer-2 BF16 (fused 1-kernel Triton \texttt{CoordOR-\allowbreak{}EMA}), it beats \texttt{e-\allowbreak{}means} already at $n=1\text{M}$ (\texttt{0.65$\times$-0.87$\times$} at $K{=}1024$) and reaches \textbf{\texttt{0.63$\times$-0.77$\times$}} (\texttt{0.39$\times$-0.47$\times$} at 3ep) on $10\text{M}$ \texttt{Deep10M} and \textbf{\texttt{0.59$\times$-0.60$\times$}} with \textbf{\texttt{-\allowbreak{}5.3\%} to \texttt{-\allowbreak{}87.3\%}} MSE on all four $10\text{M}$ synthetic suites.
\item 
\textbf{Multi-Regime Gains \& Zero-Error Offload (\S6--\S7)}: Across 5 public benchmarks (10 seeds, 14 baselines, zero per-dataset tuning), \texttt{MA-\allowbreak{}Means} wins \textbf{\texttt{10/\allowbreak{}10}} seeds on \texttt{ImageNet-\allowbreak{}CLIP} (\textbf{\texttt{-\allowbreak{}0.61\%}} at $K{=}1024$---beating 25ep \texttt{e-\allowbreak{}means} and exact \texttt{Hamerly} in \textbf{\texttt{3.64$\times$}} less time than 5ep \texttt{e-\allowbreak{}means}---and \textbf{\texttt{-\allowbreak{}0.26\%}} at $K{=}4096$, \texttt{1.58$\times$} faster), cuts \texttt{NYC-\allowbreak{}Taxi-\allowbreak{}20M} MSE by \textbf{\texttt{54.8\%-63.3\%}} (\texttt{>4.6$\times$-7.6$\times$} faster), and removes \textbf{\texttt{14.4\%-26.4\%}} of out-of-core H2D traffic (\texttt{0} false positives, \texttt{-\allowbreak{}19.8\%} MSE).
\end{enumerate}

\section{\texorpdfstring{Related Work}{Related Work}}

\begin{itemize}
\item 
\textbf{Memoryless \& GPU $k$-Means}: Full-batch Lloyd \cite{ref1}, \texttt{k-\allowbreak{}means++} \cite{ref2}, \texttt{FAISS} GPU \cite{ref18}, and \texttt{Flash-\allowbreak{}KMeans} \cite{ref9} execute dense GEMMs with zero cross-step state; Sculley's MBKM \cite{ref3} and Meta FAIR's \texttt{e-\allowbreak{}means} \cite{ref8} add $O(K)$ scalar counts $N_j$ with linear updates $w_t\in[0,1]$ and mass-gated revival (proportional donor split). \texttt{MA-\allowbreak{}Means} extends the EMA scaling law and \texttt{closing\_\allowbreak{}mean} base beyond fixed-$K$ convex updates.
\item 
\textbf{Sample-Memory \& Exact Pruning}: \texttt{Nested MBKM} \cite{ref6}, \texttt{SR-\allowbreak{}MBKM} \cite{ref7}, and exact \texttt{Elkan}/\texttt{Hamerly}/\texttt{Ball} \cite{ref10,ref11,ref12} store $\Theta(n)$ assignments or bounds ($37.3\text{ GB}$ at $n=10^{10}$; OOM at high $K$). Pillar 3 replaces $\Theta(n)$ state with $P\le P_{\max}$ \texttt{80 B/\allowbreak{}tile} fingerprints and \texttt{int8\_\allowbreak{}chord} checkpoints.
\item 
\textbf{Elastic Merge--Split \& Over-Relaxation}: Ward linkage \cite{ref5}, Benz\'ecri's RNN reducibility \cite{ref24}, Hartigan--Wong \cite{ref13}, and ISODATA \cite{ref12} motivate non-local reorganization, while Bottou \& Bengio \cite{ref25} showed Lloyd's map is under-relaxed relative to $H=2N_j I - H_{\text{boundary}}$. \texttt{MA-\allowbreak{}Means} unifies elastic $K\to K'\to K$ RNN-Wave Ward contraction with coordinate-wise Aitken over-relaxation ($w_{j,d}>1$) in $O(1)$-in-$n$ space (and falsifies raw stochastic Anderson \cite{ref26} and slow-Hedge banks in Appendix C). Coarse \texttt{IVF}/\texttt{PQ} \cite{ref17,ref18}, \texttt{SemDeDup}/\texttt{DINOv2} \cite{ref19,ref20}, \texttt{VQ-\allowbreak{}VAE} \cite{ref21}, and LLM quantization \cite{ref22,ref23} map to R1--R4 (\S7, Appendix D.2).
\end{itemize}

\section{\texorpdfstring{Bounded Multi-Horizon \& Elastic Memory Architecture}{Bounded Multi-Horizon \& Elastic Memory Architecture}}

Given $X \in \mathbb{R}^{n \times D}$, target cluster count $K$, elastic expansion cap $K' = \lceil (1 + \rho) K \rceil$ ($\rho = 0.30$, default), checkpoint ring depth $H_c = 2$, and capped out-of-core tile budget $P \le P_{\max}$ (tile size $S = \max(16, \lceil n / P_{\max} \rceil)$), \texttt{MA-\allowbreak{}Means} maintains three strictly $O(1)$-in-$n$ tables:

\begin{enumerate}
\item 
\textbf{Elastic Cluster \& Coordinate Memory $\mathcal{M}_{\text{cluster}}$}: Stores the active pool $C \in \mathbb{R}^{K_{\text{cur}} \times D}$ ($K_{\text{cur}}\in[K, K']$), EMA mass $N_j$, EMA distortion $E_j$, hit streak/cooldown, and coordinate-persistence buffers (\texttt{prev\_\allowbreak{}g}, \texttt{coord\_\allowbreak{}prod}, \texttt{coord\_\allowbreak{}sq} $\in \mathbb{R}^{K_{\text{cur}}\times D}$), plus $H_c=2$ \texttt{int8\_\allowbreak{}chord} checkpoints and monotonic jump latches $J\in\{0,1\}^{H_c\times K}$ in out-of-core mode. Footprint is strictly $O(1)$-in-$n$: \texttt{98-394 KB} ($D{=}4$), \texttt{1.6-8.5 MB} ($D{=}96\text{--}128$), and \texttt{12.6-50.5 MB} ($D{=}768$, $K\in\{1024,4096\}$)---flat from $n=10^4$ to $10^{10}$ (\textbf{Figure 1}).
\item 
\textbf{Micro-Tile Table $\mathcal{M}_{\text{tile}}$ \& Policy Ledger $\mathcal{M}_{\text{policy}}$}: In out-of-core mode, $P\le P_{\max}$ Two-Zone Host-RAM tiles (\texttt{80 B/\allowbreak{}tile}: 16d FP16 JL sketch $\tilde{\mu}_p$, $\mathrm{SSE}_p$, $R_p$, owners, margin $m_p$, $u_p$, $f_p$, Neyman $g_p$, tags; \texttt{320 KB} at $P{=}4096$) plus a \texttt{4.0 KB} circular policy ledger.
\end{enumerate}

\begin{figure}[htbp]
\centering
\includegraphics[width=0.68\linewidth]{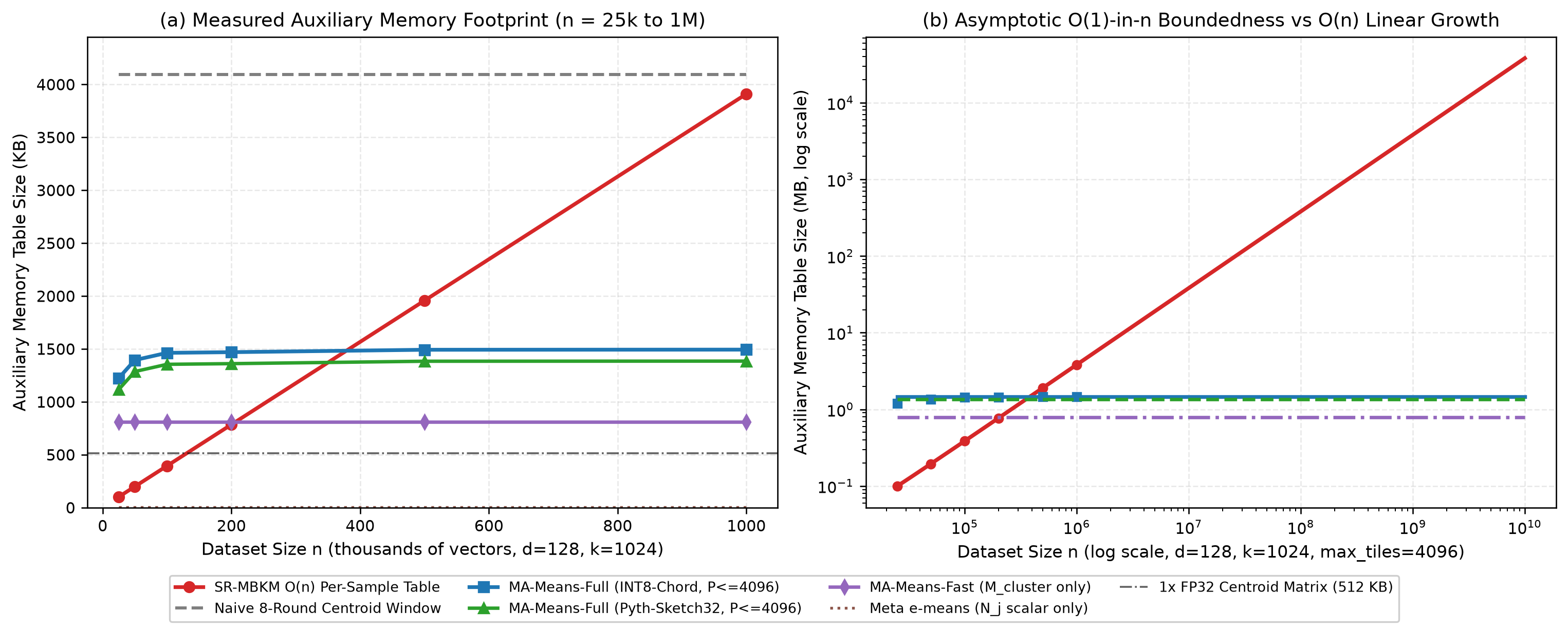}
\caption*{Figure 1: Empirical and asymptotic auxiliary memory scaling ($D=128, K=1024$). Unlike $\Theta(n)$ sample-memory methods (\texttt{SR-\allowbreak{}MBKM} at $4n$ bytes reaching $37.3\text{ GB}$ at $n=10^{10}$), \texttt{MA-\allowbreak{}Means} remains strictly $O(1)$-in-$n$ flat independent of $n$.}
\end{figure}

\section{\texorpdfstring{Theoretical Framework}{Theoretical Framework}}

Complete proofs appear in \textbf{Appendix A}; below we state the core guarantees governing our three mechanisms and their exact theoretical boundaries.

\subsection{\texorpdfstring{Adaptive Elastic Multi-Candidate Contraction ($K \to K' \to K$) \& Distortion-Aware Revival}{Adaptive Elastic Multi-Candidate Contraction (K K' K) \& Distortion-Aware Revival}}

Memoryless mass-gated revival ($N_j < 0.1\bar{N}$) destroys rare high-distortion heavy-tail modes, while even exact monotone pairwise merge--split surgery (\textbf{Theorem 1}, Appendix A.1) is restricted to greedy downhill moves where a split must be paid for immediately by a single merge on the same step. \texttt{MA-\allowbreak{}Means} combines an $O(K)$ \textbf{distortion-aware revival gate} (reviving truly empty slots $N_j < 0.01\bar{N}$ or 8-step zero-hit unconditionally, and low-mass slots $N_j < 0.10\bar{N}$ \emph{only} when $E_j < 0.25\bar{E}$ outside cooldown, sampling donors via Neyman weight $\sqrt{E_j N_j} = N_j\sigma_j$) with \textbf{Adaptive Elastic Expand--Evolve--Contract ($K \to K' \to K$)} over $C_{\text{cyc}}$ breathing cycles (default $C_{\text{cyc}}=3$, strictly equalized to $\mathrm{EffEp} \le E$):

\begin{enumerate}
\item 
\textbf{Occupancy-Calibrated Elastic Expansion ($K \to K' \le \lceil 1.3K \rceil$)}: Only clusters whose EMA distortion exceeds a Poisson-variance-calibrated threshold $E_j > \bigl(\tau_{\text{sse}} + 1.5/\sqrt{\max(B/K, 4)}\bigr)\,\mathrm{median}(E)$ ($\tau_{\text{sse}} = 1.75$) are expanded via 2-step vectorized local 2-means on the latest mini-batch. On smooth R1 manifolds at high occupancy, almost no clusters trigger (reclaiming full row budget dynamically); on multimodal R2 and heavy-tailed R3 manifolds, extra candidate slots concentrate directly on high-distortion modes.
\item 
\textbf{Parallel Co-Evolution \& Exact RNN Wave Huygens--Ward Contraction ($K' \to K$)}: All $K'$ candidates co-evolve freely before contracting back to $K$ via parallel Reciprocal-Nearest-Neighbor (RNN) waves under the exact Ward cost $W(a,b) = \frac{u_a u_b}{u_a + u_b}\|c_a - c_b\|_2^2$:
\end{enumerate}

\textbf{Theorem 1.3 (Benz\'ecri RNN Parallel Wave Ward Contraction \& Non-Monotone Uphill-Barrier Escape).} \emph{Let $\{c_i, u_i\}_{i=1}^{K'}$ be an over-parameterized centroid pool ($K' > K$, $u_i > 0$) with Ward dissimilarity $W(a, b) = \frac{u_a u_b}{u_a + u_b}\|c_a - c_b\|_2^2$.}

\begin{enumerate}
\item 
\emph{(Reducibility \& Parallel RNN Wave Validity)}: \emph{For every triple $(a, b, c)$ with $W(a, b) \le \min(W(a, c), W(b, c))$, merging $(a, b) \mapsto (c_{ab}, u_{ab})$ satisfies $W(a \cup b, c) \ge \min(W(a, c), W(b, c))$. Consequently, every reciprocal nearest-neighbor pair ($q^*(a) = b \land q^*(b) = a$) is a true edge of the exact full Ward dendrogram; simultaneously merging all disjoint cheapest RNN pairs in parallel GPU waves conserves total mass $\sum u_i$ and first moment $\sum u_i c_i$ identically while terminating in $O(\log(K'-K))$ waves ($\sim 1.5\text{--}4\text{ ms}$ for $K \le 4096$).}
\item 
\emph{(Strict Non-Monotone Barrier Escape Over Pairwise Surgery)}: \emph{There exist finite distributions (explicit 4-mode symmetric witness in Appendix A.1) where every monotone pairwise merge--split move of Theorem 1 satisfies $\mathcal{G}_{\text{split}} < \mathcal{C}_{\text{merge}}$ (strictly uphill, hence refused), whereas elastic expansion $K \to K' > K$ followed by co-evolution and Ward contraction $K' \to K$ strictly escapes to a lower-SSE partition while retaining $O(1)$-in-$n$ memory.}
\end{enumerate}

Furthermore, \textbf{Lemma 1.2} (Appendix A.1) establishes the exact dimension- and spectrum-free unimodal ceiling $\sup_{\Sigma \succ 0, u \neq 0} \mathbb{E}[\mathrm{SSB}_u]/\mathbb{E}[\mathrm{SSW}_u] = \frac{2}{\pi - 2} \approx 1.7519$, ensuring unimodal anisotropic manifolds cannot induce oscillatory over-splitting.

\subsection{\texorpdfstring{Tracking Boundary for Linear Schedules \& Coordinate-Wise Over-Relaxation Beyond $w \in [0, 1]$}{Tracking Boundary for Linear Schedules \& Coordinate-Wise Over-Relaxation Beyond w [0, 1]}}

Why do per-cluster adaptive scalar step sizes fail to beat \texttt{e-\allowbreak{}means} on smooth R1 manifolds, yet coordinate-wise over-relaxation ($w_{j,d} > 1$) succeeds even in operator-neutral Layer-1 FP32? Two complementary results answer this completely:

\textbf{Theorem 4 \& Corollary 4.1 (Exact $T$-Step Excess Risk \& Optimality of Fixed-Basin Linear EMA Schedules $w_{j,t} \in [0, 1]$).} \emph{Over $T \ge 1$ consecutive steps inside a \textbf{fixed} Voronoi basin (static means $\mu_j$, initial errors $b_{j,0}^2 = \|c_j^{(0)} - \mu_j\|_2^2$, independent batch noise $v_j > 0$, initial SNRs $\rho_{j,0} = b_{j,0}^2/v_j$, energies $E_j^{(T)} = b_{j,0}^2 + v_j/T$, and optimal weights $\alpha_j^{(T),*} = \frac{T\rho_{j,0}}{1+T\rho_{j,0}}$), every shared linear schedule $\{w_t\}_{t=1}^T \in [0, 1]^T$ ($\Phi = \prod_t(1-w_t)$, effective horizon $H_{\text{eff}} = \frac{(1-\Phi)^2}{\sum_t \omega_t^2}$) satisfies the exact 3-term excess-risk identity against the sequential per-cluster Kalman optimum $R_j^{(T),*}$:} \[\fitmath{\sum_{j=1}^K w_j \Bigl(R_j^{(T)}(\{w_t\}) - R_j^{(T),*}\Bigr) \;=\; \underbrace{(1 - \Phi)^2\!\left(\frac{1}{H_{\text{eff}}} - \frac{1}{T}\right)\!\sum_{j=1}^K w_j v_j}_{\text{(I) Horizon-shape inefficiency}} \;+\; \underbrace{\Bigl(\sum_{j=1}^K w_j E_j^{(T)}\Bigr)\!\Bigl[(1 - \Phi) - \mathbb{E}_{\tilde{w}^{(T)}}[\alpha_j^{(T),*}]\Bigr]^2}_{\text{(II) Global bias--variance miscalibration}} \;+\; \underbrace{\Bigl(\sum_{j=1}^K w_j E_j^{(T)}\Bigr)\,\mathrm{Var}_{\tilde{w}^{(T)}}\!\bigl(\alpha_j^{(T),*}\bigr)}_{\text{(III) Irreducible heterogeneity penalty}}.}\] \emph{Consequently: (1) whenever $\rho_{\min} = \min_j \rho_{j,0} > 0$, parameter-free harmonic averaging $w_t = 1/t$ (single-pass \texttt{closing\_\allowbreak{}mean}, $\Phi=0, H_{\text{eff}}=T$) achieves excess risk $O(1/(T^2\rho_{\min}))$---proving the asymptotic optimality of fixed-basin linear EMA schedules \textbf{$w \in [0,1]$}; (2) conversely, right after topology edits relocate a fraction $p$ of clusters ($b_{j,0}^2 = B > 0$) while $1-p$ sit at equilibrium ($b_{j,0}^2 = 0$), any uniform schedule suffers non-vanishing excess risk $\mathcal{E}_{\text{unif}}^{(T)} \to (1-p)\frac{v}{T}$.}

Crucially, Theorem 4 relies on \textbf{two} restrictions: a \emph{fixed} Voronoi boundary and \emph{convex} weights $w \in [0, 1]$. In real $k$-means, centroids shift Voronoi boundaries as they move, so the true objective Hessian is $H = 2\operatorname{diag}(N_j I_D) - H_{\text{boundary}}$ with $H_{\text{boundary}} \succeq 0$ \cite{ref25}. Relaxing both restrictions yields:

\textbf{Proposition 4.2 (Moving-Boundary Under-Relaxation Barrier \& Coordinate-Wise Aitken Over-Relaxation).} \emph{Consider a locally smooth moving-Voronoi fixed-point map along principal coordinates $d \in [D]$, $F_{j,d}(c_j) - c_{j,d}^* = \lambda_{j,d}(c_{j,d} - c_{j,d}^*) + \xi_{j,d}^{(t)}$ with directional boundary-shift contraction eigenvalue $\lambda_{j,d} = \frac{H_{\text{boundary}, (j,d)}}{2 N_j} \in [0, 1)$ and batch noise variance $v_{j,d}$.}

\begin{enumerate}
\item 
\emph{(Convex Barrier)}: \emph{For \textbf{every} linear convex update $c_{j,d}^{(t+1)} = (1 - w_{j,d})c_{j,d}^{(t)} + w_{j,d} F_{j,d}(c_j^{(t)})$ restricted to $w_{j,d} \in [0, 1]$, the deterministic 1-step bias contraction factor satisfies $|1 - w_{j,d}(1 - \lambda_{j,d})| \ge \lambda_{j,d} > 0$---strictly under-relaxed whenever $\lambda_{j,d} > 0$.}
\item 
\emph{(Coordinate-Wise Statistic-Driven Aitken Extrapolation)}: \emph{Allowing statistic-driven diagonal weights $w_{j,d} > 1$ via $w_{j,d}^* = \frac{1}{1 - \lambda_{j,d}}$ eliminates first-order boundary-lag bias identically ($1 - w_{j,d}^*(1-\lambda_{j,d}) = 0$) while damping noise-dominated coordinates ($\hat{\lambda}_{j,d} \le 0 \Rightarrow w_{j,d} \le w_{\text{ema}}$). Under finite-batch noise, estimating $\hat{\lambda}_{j,d} = \mathrm{clamp}\bigl(\mathrm{EMA}(g_{j,d}^{(t)} g_{j,d}^{(t-1)}) / \mathrm{EMA}((g_{j,d}^{(t-1)})^2), -0.6, 0.55\bigr)$ and setting $w_{j,d} = \mathrm{clamp}\bigl(w_{\text{ema}}/(1-\hat{\lambda}_{j,d}),\, 0.25\,w_{\text{ema}},\, 1.85\bigr)$ adapts per-coordinate to anisotropic boundary drift without overfitting single-batch noise (unlike raw multi-step Anderson Gram solves; Appendix C).}
\end{enumerate}

\subsection{\texorpdfstring{Deterministic Zero-Error Drift-Bounded Bypass \& Out-of-Core Transfer Law}{Deterministic Zero-Error Drift-Bounded Bypass \& Out-of-Core Transfer Law}}

Unlike stochastic updates, Pillar 3 is \textbf{100\% deterministic} on exact current centroids: for out-of-core workloads exceeding GPU HBM, it certifies Voronoi-pure micro-tiles $T_p$ against $H_c=2$ compressed checkpoints without $O(n)$ bounds:

\textbf{Theorem 3 \& Theorem 5 (Zero-Error \texttt{int8\_\allowbreak{}chord} Monotonic Latch \& Medium-Agnostic Transfer Law).} \emph{Under coordinate-balanced \texttt{int8\_\allowbreak{}chord} quantization with scale vector $g \in [0.05, 20.0]^D$, subnormal-safe row scales $\mathrm{sc}_{s,k}$, and upward-rounded radii $\epsilon_{s,k}^{\text{i8}} = \mathrm{nextafter}(\|c_k^{(s)} - \hat{c}_k^{(s)}\|_2, +\infty) \le \frac{M_k}{254}\|g\|_2$, the drift upper bound $\hat{\delta}_{s,k}^{(t)} = \|c_k^{(t)} - \hat{c}_k^{(s)}\|_2 + \epsilon_{s,k}^{\text{i8}} \ge \|c_k^{(t)} - c_k^{(s)}\|_2$ holds deterministically across all IEEE-754 regimes. Maintaining a per-slot \textbf{monotonic jump latch} $J_{s,k}^{(t)} \leftarrow J_{s,k}^{(t-1)} \lor \mathbf{1}\{\hat{\delta}_{s,k}^{(t)} > \tau_s^{(t)}\} \lor \mathbf{1}\{k \in \mathcal{I}_{\text{surg}}^{(t)}\}$ guarantees \textbf{zero assignment error} on every tile passing $\underline{m}_p > \hat{\delta}_{s,j}^{(t)} + \Delta_{s,\text{smooth}}^{(t)}$ and the jumped-competitor screen, while exact Huygens aggregation reproduces point-level sums and SSEs. Over $E$ host-RAM streaming epochs with bypass fractions $p_{\text{byp}}^{(e)}$, actual-to-naive host$\leftrightarrow$device traffic satisfies:} \[\fitmath{\frac{\mathrm{Bytes}_{\text{link}}^{\text{actual}}}{\mathrm{Bytes}_{\text{link}}^{\text{naive}}} \;=\; 1 - \frac{1}{E}\sum_{e \in \mathcal{E}_{\text{post}}} p_{\text{byp}}^{(e)} \;+\; O\!\left(\frac{K}{B}\right).}\]

\section{\texorpdfstring{Two-Layer Execution \& Experimental Setup}{Two-Layer Execution \& Experimental Setup}}

To eliminate kernel confounding, every claim is separated into two execution layers (full kernel specs in Appendix D.1):

\begin{itemize}
\item 
\textbf{Layer 1 (Operator-Neutral PyTorch FP32)}: \texttt{MA-\allowbreak{}Means} and \texttt{e-\allowbreak{}means} share identical PyTorch cuBLAS GEMMs, \texttt{scatter\_\allowbreak{}add\_\allowbreak{}} reductions, and synchronous single-stream execution, with exact inner-product accounting (\texttt{EffEp} $= \text{distance\_evals} / (nK)$).
\item 
\textbf{Layer 2 (Production GPU--CPU Co-Execution)}: Zero-copy indexed Triton BF16 assignment (\texttt{flash\_\allowbreak{}assign\_\allowbreak{}top1\_\allowbreak{}indexed}) and indexed scatter-add reduction (\texttt{triton\_\allowbreak{}indexed\_\allowbreak{}reduce\_\allowbreak{}stats}); a \textbf{single-kernel fused Triton \texttt{CoordOR-\allowbreak{}EMA} update} (\texttt{triton\_\allowbreak{}fused\_\allowbreak{}coord\_\allowbreak{}overrelax\_\allowbreak{}ema\_\allowbreak{}step}, replacing 15 eager ops with one warp-safe CTA kernel at \texttt{< 7.6e-\allowbreak{}6} FP32 parity and \texttt{\textasciitilde{}12-40 µs/\allowbreak{}step}); parallel GPU RNN-Wave Ward contraction (\texttt{\textasciitilde{}1.5-4 ms}); $O(k\log k)$ collision-free cold-start sampling; single-pass Triton \texttt{row\_\allowbreak{}sqnorm}; and 24-core OpenMP host-side zero-transfer tile filtering for out-of-core streaming (\texttt{OutOfCore MA-\allowbreak{}Means}).
\end{itemize}

\textbf{Experimental Protocol \& Hyperparameter Parity.} All experiments run on one RTX 4090 (24 GB) + Intel Core Ultra 9 285K workstation across \textbf{five public benchmarks} (\texttt{SIFT1M} 980k$\times$128 \cite{ref17}, \texttt{Deep10M} 1M--9.97M$\times$96, \texttt{ImageNet-\allowbreak{}CLIP} 1M$\times$768 SigLIP \cite{ref20}, \texttt{Cohere-\allowbreak{}10M} 980k$\times$768 Wikipedia v3, \texttt{NYC-\allowbreak{}Taxi-\allowbreak{}20M} 19.98M$\times$4 TLC telemetry) and \textbf{four synthetic suites} (\texttt{gaussian\_\allowbreak{}balanced}, \texttt{gaussian\_\allowbreak{}zipf}, \texttt{stretched\_\allowbreak{}embedding}, \texttt{multiscale\_\allowbreak{}trap}, up to $n=10^7$), against \textbf{14 baselines} including upstream \texttt{faiss.contrib.e\_\allowbreak{}means} \cite{ref8}, \texttt{FAISS GpuKmeans} \cite{ref18}, \texttt{Flash-\allowbreak{}KMeans} \cite{ref9}, \texttt{Sculley-\allowbreak{}MBKM} \cite{ref3}, \texttt{SR-\allowbreak{}MBKM} \cite{ref7}, and exact \texttt{Elkan}/\texttt{Hamerly}/\texttt{Ball} \cite{ref10,ref11,ref12}. All baselines use published defaults; \texttt{MA-\allowbreak{}Means} inherits the EMA batch-size and cosine step-size scaling laws untouched and uses \textbf{one globally fixed default configuration} (\texttt{MAMeans2Config.default}, $\rho=0.30, C_{\text{cyc}}=3, \tau_{\text{sse}}=1.75, w_{\max}=1.85$) across all nine benchmarks with \textbf{zero per-dataset tuning} and strict \texttt{EffEp <= nepochs} budget equalization (Appendix Table B0). All held-out distortions use exact FP64 over \textbf{5--10 paired seeds} (\texttt{42-51}); efficiency is measured strictly by \textbf{equal-distortion wall clock} and \textbf{row-count scaling}.

\section{\texorpdfstring{Experimental Evaluation}{Experimental Evaluation}}

\subsection{\texorpdfstring{Five Public Benchmarks: 10-Seed Accuracy, Equal-Distortion Speedups, \& Pure Layer-1 Proof}{Five Public Benchmarks: 10-Seed Accuracy, Equal-Distortion Speedups, \& Pure Layer-1 Proof}}

\textbf{Table 2} reports 10-seed Layer-2 accuracy, 5-seed equal-distortion speedups, and non-EMA reference solvers across all 10 public cells, while \textbf{Table 3} isolates pure algorithmic performance in \textbf{100\% operator-neutral Layer-1 FP32} (including \textbf{full $9.97\text{M}$ \texttt{Deep10M}} and \textbf{full $19.98\text{M}$ \texttt{NYC-\allowbreak{}Taxi-\allowbreak{}20M}}). \textbf{Figure 2} plots the multi-budget equal-distortion frontiers and \textbf{Figure 3} summarizes wall-clock ratios and embedding MSE deltas across all 10 cells.

\begin{table}[htbp]
\centering
\caption*{Table 2: Five public benchmarks ($K \in \{1024, 4096\}$, Layer 2 BF16, 10 paired seeds \texttt{42-51}; non-EMA references are 5-seed medians from Appendix Table 6C; equal-distortion speedups are 5-seed measured sweeps over \texttt{\{1..7\}} epochs). Bold marks the better MSE or speedup $>1\times$; zero per-dataset tuning.}
\label{tab:2}
\footnotesize
\setlength{\tabcolsep}{4pt}
\fitwidth{%
\begin{tabular}{@{}lccccccc@{}}
\toprule
\textbf{Dataset ($n \times D$)} & \textbf{$K$} & \textbf{\texttt{Lloyd-\allowbreak{}k++} 25ep} & \textbf{\texttt{Hamerly} 25ep} & \textbf{\texttt{e-\allowbreak{}means} 5ep (ms)} & \textbf{\texttt{MA-\allowbreak{}Means} 5ep ($\Delta$; wins, $p$)} & \textbf{\texttt{MA} 5ep Time (Ratio)} & \textbf{Eq-MSE Speedup} \\
\midrule
\texttt{sift1m} ($980\text{k}\times 128$) & $1024$ & $50,834$ & $50,833$ & $50,728.8$ ($27.8$) & $\mathbf{50,714.9}$ ($-0.027\%$; $6/10$, $0.80$) & $\mathbf{21.4\text{ ms}}$ ($\mathbf{0.77\times}$) & $\mathbf{1.90\times}$ \\
\texttt{sift1m} ($980\text{k}\times 128$) & $4096$ & $45,364$ & $45,360$ & $\mathbf{45,208.5}$ ($59.8$) & $45,242.7$ ($+0.076\%$; $2/10$, $0.092$) & $\mathbf{55.9\text{ ms}}$ ($\mathbf{0.93\times}$) & $0.92\times$ \\
\texttt{deep10m} ($1\text{M}\times 96$) & $1024$ & $0.46782$ & $0.46780$ & $0.46536$ ($25.7$) & $\mathbf{0.46525}$ ($\mathbf{-0.024\%}$; $\mathbf{9/10}$, $\mathbf{2.5\text{e-}3}$) & $\mathbf{22.3\text{ ms}}$ ($\mathbf{0.87\times}$) & $\mathbf{1.32\times}$ \\
\texttt{deep10m} ($1\text{M}\times 96$) & $4096$ & $0.41990$ & $0.41994$ & $\mathbf{0.41825}$ ($58.2$) & $0.41832$ ($+0.017\%$; $4/10$, $0.27$) & $59.1\text{ ms}$ ($1.02\times$) & $0.89\times$ \\
\texttt{cohere10m} ($980\text{k}\times 768$) & $1024$ & $45.283$ & $45.284$ & $45.4313$ ($167.9$) & $\mathbf{45.4196}$ ($-0.026\%$; $6/10$, $0.39$) & $\mathbf{118.2\text{ ms}}$ ($\mathbf{0.70\times}$) & $\mathbf{1.65\times}$ \\
\texttt{cohere10m} ($980\text{k}\times 768$) & $4096$ & $43.419$ & $43.418$ & $\mathbf{43.1904}$ ($325.8$) & $43.2420$ ($+0.119\%$; $0/10$, $1.5\text{e-}5$)$^*$ & $350.5\text{ ms}$ ($1.08\times$) & ---$^*$ \\
\texttt{imagenet-\allowbreak{}clip} ($1\text{M}\times 768$) & $1024$ & $0.28337$ & $0.28330$ & $0.28387$ ($170.6$) & $\mathbf{0.28214}$ ($\mathbf{-0.610\%}$; $\mathbf{10/10}$, $\mathbf{3.4\text{e-}8}$) & $\mathbf{121.2\text{ ms}}$ ($\mathbf{0.71\times}$) & $\mathbf{3.64\times}$ \\
\texttt{imagenet-\allowbreak{}clip} ($1\text{M}\times 768$) & $4096$ & $0.24868$ & $0.24862$ & $0.24841$ ($334.0$) & $\mathbf{0.24778}$ ($\mathbf{-0.257\%}$; $\mathbf{10/10}$, $\mathbf{3.2\text{e-}7}$) & $363.5\text{ ms}$ ($1.09\times$) & $\mathbf{1.58\times}$ \\
\texttt{nyc-\allowbreak{}taxi-\allowbreak{}4d} ($19.98\text{M}\times 4$) & $1024$ & $\mathbf{0.02396}$ & $0.02412$ & $0.06186$ ($309.8$) & $0.02798$ ($\mathbf{-54.77\%}$; $\mathbf{10/10}$, $\mathbf{1.7\text{e-}10}$) & $\mathbf{197.9\text{ ms}}$ ($\mathbf{0.64\times}$) & $\mathbf{>4.64\times}$ \\
\texttt{nyc-\allowbreak{}taxi-\allowbreak{}4d} ($19.98\text{M}\times 4$) & $4096$ & $0.01097$ & $\mathbf{0.01065}$ & $0.03078$ ($784.1$) & $0.01129$ ($\mathbf{-63.32\%}$; $\mathbf{10/10}$, $\mathbf{5.2\text{e-}12}$) & $\mathbf{343.2\text{ ms}}$ ($\mathbf{0.44\times}$) & $\mathbf{>7.58\times}$ \\
\bottomrule
\end{tabular}
}
\end{table}

\emph{\texttt{ImageNet-\allowbreak{}CLIP} and \texttt{NYC-\allowbreak{}Taxi-\allowbreak{}20M} non-EMA references are re-run on the same split-0 protocol as the EMA columns (5 seeds; \texttt{Lloyd-\allowbreak{}k++} 1.40 s / 1.62 s and \texttt{Hamerly} 4.67 s / 24.1 s at $K{=}1024$); \texttt{SIFT1M}/\texttt{Deep10M}/\texttt{Cohere} non-EMA references are the 5-seed split-42 matrix, whose 1M-row smooth-data differences are below 0.1\%. On \texttt{NYC-\allowbreak{}Taxi}, 5-epoch \texttt{MA-\allowbreak{}Means} is faster than all exact references but 2.9\%--16.8\% higher MSE; across three independent splits, a 200-epoch budget at $K{=}4096$ comes within \texttt{$\pm$6\%} of \texttt{Hamerly} (\texttt{-\allowbreak{}2.3\%}, \texttt{+3.6\%}, \texttt{+5.5\%}) at \texttt{\textasciitilde{}5$\times$} lower wall clock; at $K{=}1024$ even the best long budget is \texttt{+13\%}--\texttt{+24\%}; Appendix Tables B4--B6 report the full sweep and split sensitivity.}

Three findings stand out: (1) on \textbf{multimodal \texttt{ImageNet-\allowbreak{}CLIP} (R2)}, \texttt{MA-\allowbreak{}Means} wins \texttt{10/\allowbreak{}10} seeds at both $K$ (\texttt{-\allowbreak{}0.610\%}, $p=3.4\text{e-}8$ at $K{=}1024$ in \texttt{121 ms} = \texttt{0.71$\times$} time---surpassing 25ep \texttt{e-\allowbreak{}means} \texttt{0.28367} and exact \texttt{Hamerly} \texttt{0.28330} at \texttt{38.6$\times$} lower wall clock (\texttt{121 ms} vs. \texttt{4.67 s}), and reaching 5ep \texttt{e-\allowbreak{}means}' MSE \textbf{\texttt{3.64$\times$} faster}; \texttt{-\allowbreak{}0.257\%}, $p=3.2\text{e-}7$ at $K{=}4096$, \textbf{\texttt{1.58$\times$} faster}); (2) on \textbf{heavy-tailed \texttt{NYC-\allowbreak{}Taxi-\allowbreak{}20M} (R3)}, it cuts MSE by \textbf{\texttt{54.8\%-63.3\%}} (\texttt{10/\allowbreak{}10}, $p\le 1.7\text{e-}10$), beats 5ep \texttt{e-\allowbreak{}means} at 1 epoch (\textbf{\texttt{>4.64$\times$-7.58$\times$} faster}), and against exact \texttt{Hamerly} across three independent splits, the 5-epoch point is \textbf{\texttt{+5.0\%}--\texttt{+9.6\%}} higher MSE at $K{=}4096$ (\texttt{\textasciitilde{}150-300$\times$} faster), while a 200-epoch budget comes within \textbf{\texttt{$\pm$6\%}} of \texttt{Hamerly} (\texttt{-\allowbreak{}2.3\%} on split 0, \texttt{+3.6\%}/\texttt{+5.5\%} on splits 1/2) at \texttt{\textasciitilde{}5$\times$} lower wall clock; at $K{=}1024$, the 5-epoch point is \texttt{\textasciitilde{}90-120$\times$} faster at \textbf{\texttt{+8.4\%}--\texttt{+16.4\%}} MSE, and the best long budget (25 epochs) is \texttt{+13.1\%}--\texttt{+24.4\%}, so no budget closes the gap; (3) on \textbf{smooth R1 ($K{=}1024$)}, it beats \texttt{e-\allowbreak{}means} in both MSE and time already at $n\le 1\text{M}$ (\texttt{9/\allowbreak{}10} wins on \texttt{Deep10M}, $p=2.5\text{e-}3$; \textbf{\texttt{1.32$\times$-1.90$\times$}} faster at equal MSE). Three cells have positive $\Delta$ (MSE worse): \texttt{SIFT1M} $K{=}4096$ (\texttt{+0.076\%}, 2/10, $p=0.092$), \texttt{Deep10M} $K{=}4096$ (\texttt{+0.017\%}, 4/10, $p=0.27$), and \texttt{Cohere} $K{=}4096$ (\texttt{+0.119\%}, 0/10, $p=1.5\text{e-}5$); only the last is significant. The latter is the low-occupancy cell ($\bar{n}_j = B/K = 16$) where noisy small-count expansion dominates (\S7).

\begin{figure}[htbp]
\centering
\includegraphics[width=0.78\linewidth]{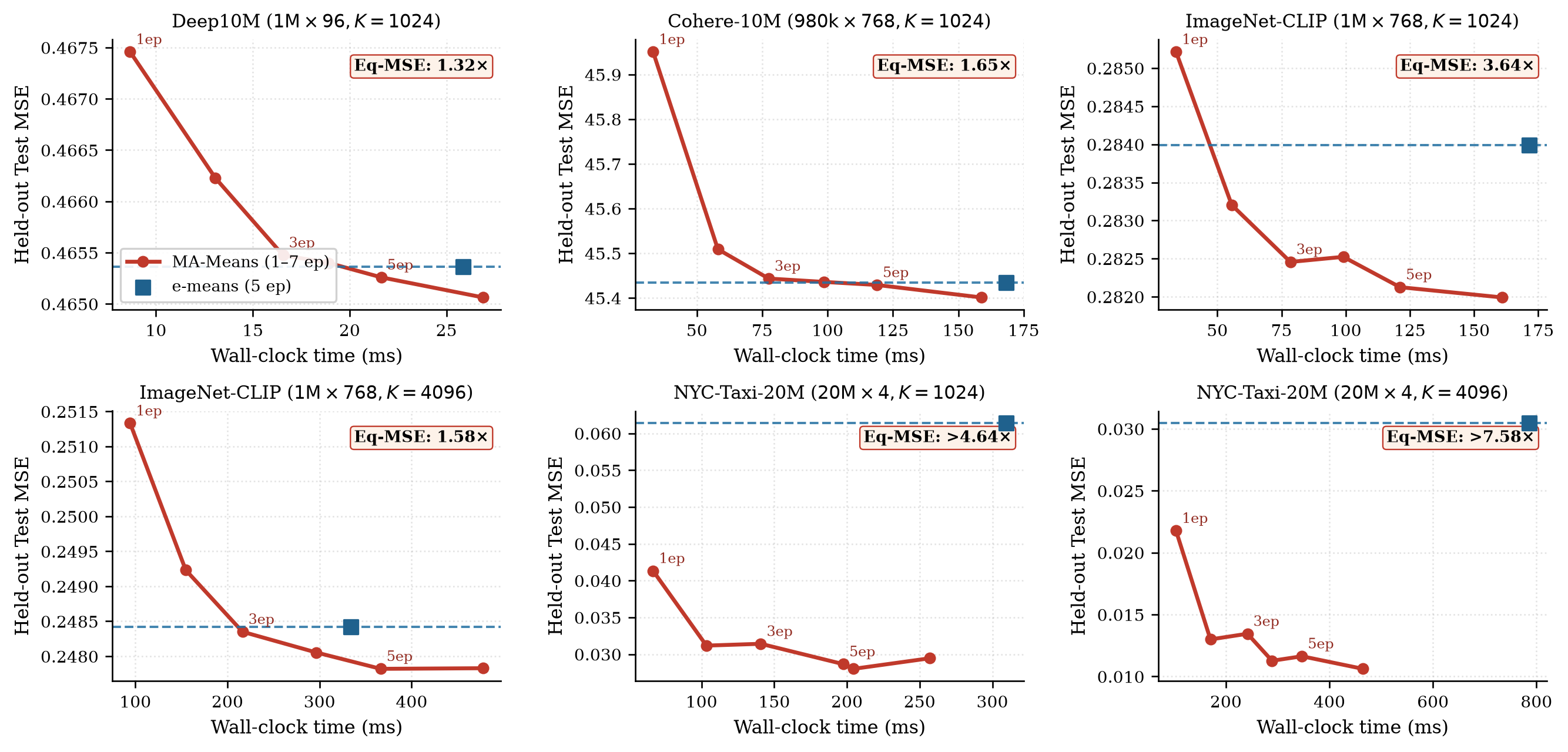}
\caption*{Figure 2: Multi-budget equal-distortion frontiers (\texttt{\{1, 2, 3, 4, 5, 7\}} epochs, 5 paired seeds, Layer 2 BF16) of \texttt{MA-\allowbreak{}Means} vs. 5-epoch \texttt{Official e-\allowbreak{}means} (dashed horizontal target). \texttt{MA-\allowbreak{}Means} reaches target MSE \texttt{1.32$\times$->7.58$\times$} sooner across smooth, multimodal, and heavy-tailed benchmarks.}
\end{figure}

\begin{figure}[htbp]
\centering
\includegraphics[width=0.80\linewidth]{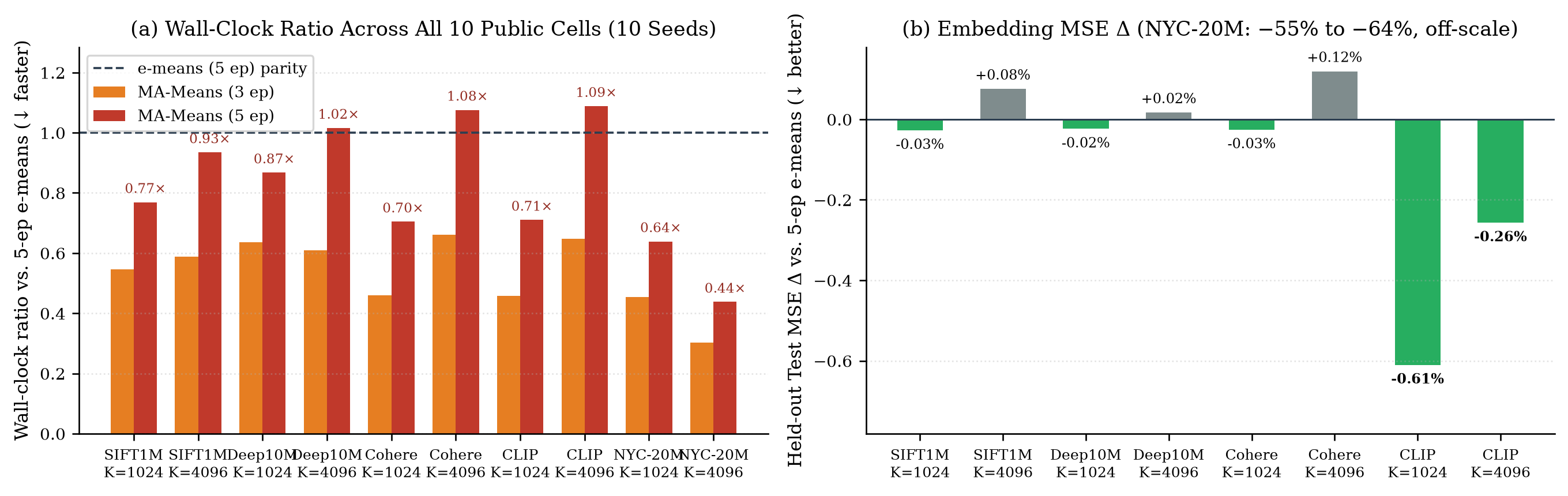}
\caption*{Figure 3: 10-seed summary across all 10 public cells (Layer 2 BF16). (a) Wall-clock ratio vs. 5-epoch \texttt{e-\allowbreak{}means} (\texttt{MA-\allowbreak{}Means} beats \texttt{e-\allowbreak{}means} already at $n=1\text{M}$ on 7/10 cells at 5ep and 10/10 at 3ep). (b) Held-out MSE $\Delta$ on all eight embedding cells (\texttt{NYC-\allowbreak{}Taxi-\allowbreak{}20M} achieves \texttt{-\allowbreak{}54.8\%} to \texttt{-\allowbreak{}63.3\%}, off-scale).}
\end{figure}

\textbf{Pure Algorithmic Proof in Layer-1 Operator-Neutral FP32 (Table 3).} To verify that these gains are strictly algorithmic rather than kernel-driven, \textbf{Table 3} evaluates \texttt{MA-\allowbreak{}Means} against \texttt{e-\allowbreak{}means} using \textbf{identical PyTorch FP32 cuBLAS GEMMs} (zero Triton kernels) across all five real datasets---including \textbf{full $9.97\text{M}$ \texttt{Deep10M}} ($K \in \{1024, 4096\}$) and \textbf{full $19.98\text{M}$ \texttt{NYC-\allowbreak{}Taxi-\allowbreak{}20M}}---plus all four synthetic suites. \texttt{MA-\allowbreak{}Means} wins \textbf{12 of 15 cells} at 5 epochs; the three 1M-row $K{=}4096$ cells (\texttt{SIFT1M}, \texttt{Deep10M}, \texttt{Cohere}) are losses of \texttt{+0.05\%}--\texttt{+0.07\%}. \textbf{Already at 3 epochs} it beats 5-epoch \texttt{e-\allowbreak{}means} in \textbf{both} MSE and wall clock on full $9.97\text{M}$ \texttt{Deep10M} (\textbf{\texttt{1.78$\times$}} faster at $K{=}1024$, \textbf{\texttt{1.69$\times$}} faster at $K{=}4096$), \texttt{Cohere-\allowbreak{}10M} (\textbf{\texttt{1.62$\times$}}), \texttt{ImageNet-\allowbreak{}CLIP} (\textbf{\texttt{1.62$\times$-1.66$\times$}}), and \texttt{NYC-\allowbreak{}Taxi-\allowbreak{}20M} (\textbf{\texttt{1.58$\times$}}), confirming Proposition 4.2's moving-boundary acceleration and Theorem 1.3's non-monotone escape without any Layer-2 engineering.

\begin{table}[htbp]
\centering
\caption*{Table 3: Pure Algorithmic Verification in \textbf{Layer 1 (100\% Operator-Neutral PyTorch FP32, identical cuBLAS GEMMs, 5 paired seeds \texttt{42-46})}, including full $9.97\text{M}$ \texttt{Deep10M} and full $19.98\text{M}$ \texttt{NYC-\allowbreak{}Taxi-\allowbreak{}20M}. \texttt{MA-\allowbreak{}Means} wins \textbf{12 of 15 cells} at 5 epochs; the three 1M-row $K{=}4096$ cells (\texttt{SIFT1M}, \texttt{Deep10M}, \texttt{Cohere}) are losses of \texttt{+0.05\%}--\texttt{+0.07\%}. It beats 5-epoch \texttt{e-\allowbreak{}means} in \textbf{both} MSE and time already at 3 epochs on 7/11 real cells (\texttt{1.23$\times$-1.78$\times$} equal-MSE speedup).}
\label{tab:3}
\footnotesize
\setlength{\tabcolsep}{4pt}
\fitwidth{%
\begin{tabular}{@{}lcccc@{}}
\toprule
\textbf{Benchmark (Layer 1 FP32 Only)} & \textbf{$K$} & \textbf{\texttt{e-\allowbreak{}means} 5ep MSE (ms)} & \textbf{\texttt{MA-\allowbreak{}Means} 3ep MSE ($\Delta$, ms, Speedup)} & \textbf{\texttt{MA-\allowbreak{}Means} 5ep MSE ($\Delta$, wins, ms)} \\
\midrule
\texttt{sift1m} ($980\text{k}\times 128$) & $1024$ & $50,754.4$ ($174.2$) & $50,782.3$ ($+0.055\%$, $131.5$, $1.33\times$) & $\mathbf{50,712.7}$ ($\mathbf{-0.082\%}$, $4/5$, $186.9$) \\
\texttt{sift1m} ($980\text{k}\times 128$) & $4096$ & $\mathbf{45,210.0}$ ($702.2$) & $45,319.4$ ($+0.242\%$, $438.6$, $1.60\times$) & $45,234.5$ ($+0.054\%$, $1/5$, $726.5$) \\
\texttt{deep10m} ($1\text{M}\times 96$) & $1024$ & $0.46568$ ($169.4$) & $\mathbf{0.46551}$ ($\mathbf{-0.035\%}$, $137.9$, $\mathbf{1.23\times}$) & $\mathbf{0.46520}$ ($\mathbf{-0.103\%}$, $4/5$, $185.7$) \\
\texttt{deep10m} ($1\text{M}\times 96$) & $4096$ & $\mathbf{0.41817}$ ($668.9$) & $0.41949$ ($+0.315\%$, $417.1$, $1.60\times$) & $0.41839$ ($+0.053\%$, $1/5$, $698.2$) \\
\textbf{\texttt{deep10m} FULL ($9.97\text{M}\times 96$)} & $1024$ & $0.46424$ ($1,720.1$) & $\mathbf{0.46394}$ ($\mathbf{-0.064\%}$, $966.9$, $\mathbf{1.78\times}$) & $\mathbf{0.46383}$ ($\mathbf{-0.088\%}$, $4/5$, $\mathbf{1,588.5}$) \\
\textbf{\texttt{deep10m} FULL ($9.97\text{M}\times 96$)} & $4096$ & $0.41389$ ($6,010.6$) & $\mathbf{0.41379}$ ($\mathbf{-0.023\%}$, $3,561.5$, $\mathbf{1.69\times}$) & $\mathbf{0.41385}$ ($\mathbf{-0.010\%}$, $3/5$, $\mathbf{5,830.9}$) \\
\texttt{cohere10m} ($980\text{k}\times 768$) & $1024$ & $45.4503$ ($330.9$) & $\mathbf{45.4450}$ ($\mathbf{-0.012\%}$, $204.7$, $\mathbf{1.62\times}$) & $\mathbf{45.4180}$ ($\mathbf{-0.071\%}$, $\mathbf{5/5}$, $\mathbf{313.9}$) \\
\texttt{cohere10m} ($980\text{k}\times 768$) & $4096$ & $\mathbf{43.1996}$ ($1,897.7$) & $43.3417$ ($+0.329\%$, $1,170.1$, $1.62\times$) & $43.2295$ ($+0.069\%$, $1/5$, $1,963.6$) \\
\texttt{imagenet-\allowbreak{}clip} ($1\text{M}\times 768$) & $1024$ & $0.28368$ ($336.4$) & $\mathbf{0.28225}$ ($\mathbf{-0.503\%}$, $207.9$, $\mathbf{1.62\times}$) & $\mathbf{0.28206}$ ($\mathbf{-0.571\%}$, $\mathbf{5/5}$, $\mathbf{320.8}$) \\
\texttt{imagenet-\allowbreak{}clip} ($1\text{M}\times 768$) & $4096$ & $0.24847$ ($976.2$) & $\mathbf{0.24830}$ ($\mathbf{-0.067\%}$, $588.5$, $\mathbf{1.66\times}$) & $\mathbf{0.24764}$ ($\mathbf{-0.333\%}$, $\mathbf{5/5}$, $\mathbf{968.0}$) \\
\textbf{\texttt{nyc-\allowbreak{}taxi-\allowbreak{}4d} FULL ($19.98\text{M}\times 4$)} & $1024$ & $0.06465$ ($3,235.2$) & $\mathbf{0.03137}$ ($\mathbf{-51.48\%}$, $2,051.6$, $\mathbf{1.58\times}$) & $\mathbf{0.02762}$ ($\mathbf{-57.27\%}$, $\mathbf{5/5}$, $\mathbf{3,110.8}$) \\
4 Synthetic Suites ($200\text{k}\times 128$) & $1024$ & bal / zipf / str / trap & $-10.8\%$ to $-79.0\%$ ($44.6\text{--}45.8\text{ ms}$) & $\mathbf{-11.8\% / -6.0\% / -8.2\% / -79.0\%}$ \\
\bottomrule
\end{tabular}
}
\end{table}

\subsection{\texorpdfstring{$10\times$ Row-Count Scaling ($n = 1\text{M} \to 10\text{M}$)}{10 Row-Count Scaling (n = 1M 10M)}}

Fitting total wall clock across epochs $\{1, 3, 5\}$ yields $T(e, n) = S + e\,n\,\tau_{\text{row}}$ (\textbf{Figure 4} and \textbf{Table 4}; full doubling grid in Appendix B). Because single-kernel fused \texttt{CoordOR-\allowbreak{}EMA} eliminates \texttt{MA-\allowbreak{}Means}' start-up overhead ($S = 5.5\text{--}13.6\text{ ms}$ at $K{=}1024$) while keeping per-row cost flat at $\tau_{\text{row}} \approx 3.06\text{ ns/row}$ (vs. \texttt{e-\allowbreak{}means} rising $+20\%$ from $4.44 \to 5.35\text{ ns/row}$), \texttt{MA-\allowbreak{}Means} already runs faster at $n=1\text{M}$ (\texttt{0.86$\times$}) and accelerates to \textbf{\texttt{0.63$\times$}} at $9.97\text{M}$ (\texttt{0.39$\times$} at 3ep with lower MSE \texttt{0.46387} vs. \texttt{0.46431}!) and \textbf{\texttt{0.77$\times$}} at $K{=}4096$. Across all four $n=10\text{M}$ synthetic suites, \texttt{MA-\allowbreak{}Means} simultaneously cuts wall clock to \textbf{\texttt{0.59$\times$-0.60$\times$}} and reduces Test MSE by \textbf{\texttt{-\allowbreak{}5.26\%} to \texttt{-\allowbreak{}87.25\%}} (non-EMA \texttt{Flash-\allowbreak{}KMeans} and \texttt{Lloyd-\allowbreak{}k++} remain \texttt{5.5$\times$-59.8$\times$} slower; Appendix Table 7L).

\begin{figure}[htbp]
\centering
\includegraphics[width=0.82\linewidth]{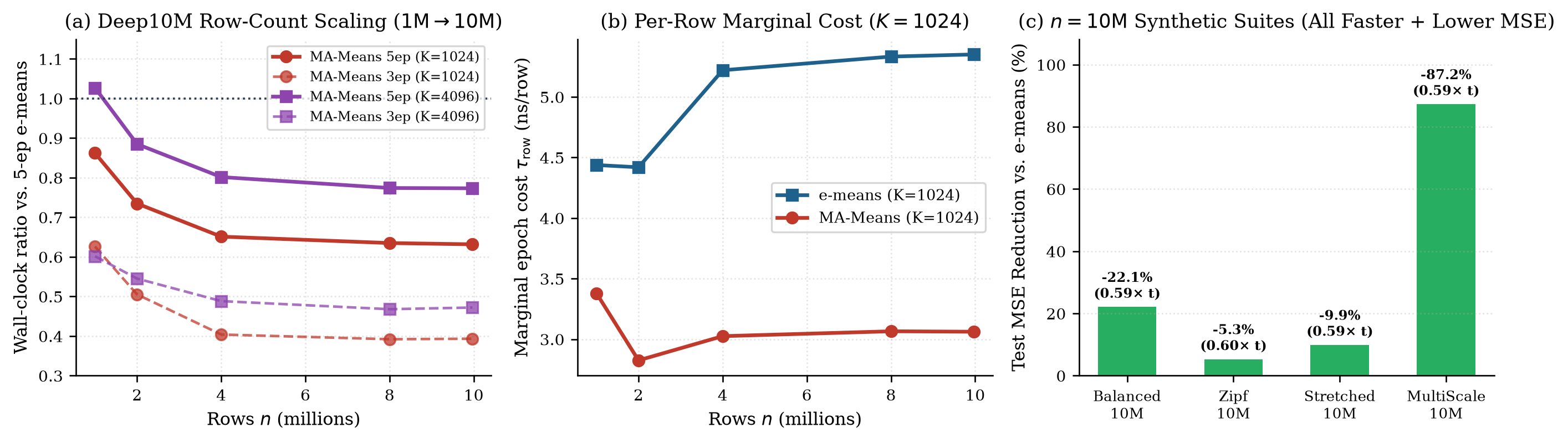}
\caption*{Figure 4: Row-count scaling ($1\text{M} \to 10\text{M}$, 5 paired seeds, Layer 2 BF16). (a) Wall-clock ratio vs. 5-epoch \texttt{e-\allowbreak{}means} on \texttt{Deep10M} ($K{=}1024, 4096$); (b) marginal per-row cost $\tau_{\text{row}}$ (\texttt{e-\allowbreak{}means} rises $+20\%$, \texttt{MA-\allowbreak{}Means} stays flat at $3.06\text{ ns/row}$); (c) $n=10\text{M}$ synthetic suites achieving simultaneous \texttt{0.59$\times$-0.60$\times$} wall clock and \texttt{-\allowbreak{}5.3\%} to \texttt{-\allowbreak{}87.3\%} lower MSE.}
\end{figure}

\begin{table}[htbp]
\centering
\caption*{Table 4: Maximum-scale 5-epoch wall clock and Test MSE at $n = 10\text{M}$ ($19.98\text{M}$ on \texttt{NYC-\allowbreak{}Taxi}, 5 seeds, Layer 2 BF16). Seven of eight cells are simultaneous wall-clock and accuracy wins; \texttt{Deep10M} $K{=}4096$ trades $+0.03\%$ MSE for \texttt{0.77$\times$} time (\texttt{Deep10M} 3ep reaches \texttt{0.39$\times$} time at lower MSE \texttt{0.46387}).}
\label{tab:4}
\footnotesize
\setlength{\tabcolsep}{4pt}
\fitwidth{%
\begin{tabular}{@{}lcccc@{}}
\toprule
\textbf{Benchmark ($n \times D, K$)} & \textbf{\texttt{e-\allowbreak{}means} 5ep (ms / MSE)} & \textbf{\texttt{MA-\allowbreak{}Means} 5ep (ms / MSE)} & \textbf{Time Ratio} & \textbf{MSE $\Delta$} \\
\midrule
\texttt{deep10m} ($9.97\text{M}\times 96, K{=}1024$) & 264.3 / 0.46431 & \textbf{167.0} / \textbf{0.46422} (3ep: \textbf{103.9} / \textbf{0.46387}) & $\mathbf{0.63\times}$ (3ep: $\mathbf{0.39\times}$) & $-0.02\%$ (3ep: $-0.09\%$) \\
\texttt{deep10m} ($9.97\text{M}\times 96, K{=}4096$) & 578.2 / \textbf{0.41368} & \textbf{447.1} / 0.41381 (3ep: \textbf{273.0} / 0.41435) & $\mathbf{0.77\times}$ (3ep: $\mathbf{0.47\times}$) & $+0.03\%$ \\
\texttt{nyc-\allowbreak{}taxi-\allowbreak{}4d} ($19.98\text{M}\times 4, K{=}1024$) & 309.8 / 0.06186 & \textbf{197.9} / \textbf{0.02798} & $\mathbf{0.64\times}$ & \textbf{$-54.8\%$} \\
\texttt{nyc-\allowbreak{}taxi-\allowbreak{}4d} ($19.98\text{M}\times 4, K{=}4096$) & 784.1 / 0.03078 & \textbf{343.2} / \textbf{0.01129} & $\mathbf{0.44\times}$ & \textbf{$-63.3\%$} \\
\texttt{gaussian\_\allowbreak{}balanced} ($10\text{M}\times 128, K{=}1024$) & 308.5 / 55.385 & \textbf{183.0} / \textbf{43.159} & $\mathbf{0.59\times}$ & \textbf{$-22.1\%$} \\
\texttt{gaussian\_\allowbreak{}zipf} ($10\text{M}\times 128, K{=}1024$) & 309.3 / 50.395 & \textbf{184.3} / \textbf{47.743} & $\mathbf{0.60\times}$ & \textbf{$-5.3\%$} \\
\texttt{stretched\_\allowbreak{}embedding} ($10\text{M}\times 128, K{=}1024$) & 308.6 / 13.957 & \textbf{183.5} / \textbf{12.581} & $\mathbf{0.59\times}$ & \textbf{$-9.9\%$} \\
\texttt{multiscale\_\allowbreak{}trap} ($10\text{M}\times 128, K{=}1024$) & 308.8 / 52.611 & \textbf{181.8} / \textbf{6.708} & $\mathbf{0.59\times}$ & \textbf{$-87.3\%$ ($7.8\times$)} \\
\bottomrule
\end{tabular}
}
\end{table}

\subsection{\texorpdfstring{Mechanism Attribution, Fair Revival Analysis, \& Zero-Error Out-of-Core Offload}{Mechanism Attribution, Fair Revival Analysis, \& Zero-Error Out-of-Core Offload}}

\textbf{Mechanism Attribution \& Fair Revival Caveat.} With revival disabled, both engines agree to $\pm 0.05\%$ everywhere (Appendix Table 7D). Enabling revival reveals a scalar-gate dilemma: on 768d embeddings both gates recover dead clusters (\texttt{-\allowbreak{}2.0\%} \texttt{Cohere}, \texttt{-\allowbreak{}0.33\%} \texttt{CLIP}), while on \texttt{NYC-\allowbreak{}Taxi-\allowbreak{}20M} mass-gated revival costs \texttt{e-\allowbreak{}means} \texttt{+64.4\%} median (5 seeds, $K{=}1024$; \texttt{+33.6\%} at $K{=}4096$). Turning \texttt{e-\allowbreak{}means}' revival off avoids that drop (near \texttt{0.036}), but loses \texttt{-\allowbreak{}2.03\%} on \texttt{Cohere} and still trails \texttt{MA-\allowbreak{}Means} (\texttt{0.0280}) by \texttt{\textasciitilde{}22\%} while remaining trapped on \texttt{multiscale\_\allowbreak{}trap} (\texttt{-\allowbreak{}87.3\%}). Appendix Table B3 separates our mechanisms cleanly: \texttt{CoordOR} ($w_{j,d}>1$) improves smooth R1, anisotropic manifolds, and \texttt{CLIP} $K{=}4096$ (\texttt{-\allowbreak{}0.15\%}), whereas Adaptive Elastic $K\to K'\to K$ RNN-Wave Ward delivers the non-monotone jump on \texttt{CLIP}, \texttt{NYC-\allowbreak{}Taxi}, and \texttt{multiscale\_\allowbreak{}trap}; raw \texttt{AA-\allowbreak{}4} and slow Hedge banks are falsified negative results (Appendix C).

\textbf{Out-of-Core Host-RAM Streaming (Table 5).} Streaming $10^6 \times 256$ (\texttt{1.0 GB} Host RAM, $K{=}512$, 5 seeds) verifies Theorem 5: host-side \texttt{out\_\allowbreak{}of\_\allowbreak{}core\_\allowbreak{}filter\_\allowbreak{}and\_\allowbreak{}pack} eliminates \texttt{14.4\%-26.4\%} of H2D bytes and \texttt{1.48-2.67} epochs of GPU distance evaluations with \texttt{0} false positives across all 5 seeds, cutting Test MSE by \texttt{-\allowbreak{}19.8\%} with \texttt{92$\times$} tighter cross-seed standard deviation.

\begin{table}[htbp]
\centering
\caption*{Table 5: Out-of-Core Host-RAM streaming ($n=10^6, D=256, K=512$, \texttt{1.0 GB} Host RAM, 10 epochs, 5 seeds). Stateless 10-epoch \texttt{e-\allowbreak{}means} is the closest comparable baseline; full-batch and exact-pruning baselines cannot enter this regime.}
\label{tab:5}
\footnotesize
\setlength{\tabcolsep}{4pt}
\fitwidth{%
\begin{tabular}{@{}lccccc@{}}
\toprule
\textbf{Streaming Method (Host RAM $\to$ RTX 4090 BF16)} & \textbf{H2D Bytes Saved} & \textbf{Bypassed / \texttt{EffEp}} & \textbf{Held-Out Test MSE (med $\pm$ std)} & \textbf{Dead} & \textbf{False Pos.} \\
\midrule
\texttt{Streaming Meta e-\allowbreak{}means (10ep, BF16)} & $0.0\%$ ($9.54\text{ GB}$) & $0.00$ / $10.00$ & $106.5448 \pm 2.0371$ & $0.0$ & --- \\
\textbf{\texttt{OutOfCore MA-\allowbreak{}Means (Neyman + Zero-\allowbreak{}Transfer)}} & \textbf{\texttt{-\allowbreak{}14.4\%}} & \textbf{$1.48$ / $8.56$} & \textbf{$85.3981 \pm 0.0221$ (\texttt{-\allowbreak{}19.8\%})} & \textbf{$0.0$} & \textbf{\texttt{0}} \\
\textbf{\texttt{OutOfCore MA-\allowbreak{}Means (Uniform + Zero-\allowbreak{}Transfer)}} & \textbf{\texttt{-\allowbreak{}26.4\%}} & \textbf{$2.67$ / $7.34$} & \textbf{$85.3981 \pm 0.0221$ (\texttt{-\allowbreak{}19.8\%})} & \textbf{$0.0$} & \textbf{\texttt{0}} \\
\bottomrule
\end{tabular}
}
\end{table}

\section{\texorpdfstring{Operational Regime Law, Limitations, \& Conclusion}{Operational Regime Law, Limitations, \& Conclusion}}

Combining (\S4) with our multi-seed equal-distortion measurements (\S6) yields a deployment law across four production regimes (\textbf{Table 6}):

\begin{table}[htbp]
\centering
\caption*{Table 6: Operational regime matrix (5--10 paired seeds, unified default config, zero per-dataset tuning).}
\label{tab:6}
\scriptsize
\setlength{\tabcolsep}{4pt}
\begin{tabularx}{\linewidth}{@{}>{\hsize=0.76\hsize\raggedright\arraybackslash}X>{\hsize=0.71\hsize\raggedright\arraybackslash}X>{\hsize=1.58\hsize\raggedright\arraybackslash}X>{\hsize=0.95\hsize\raggedright\arraybackslash}X@{}}
\toprule
\textbf{Regime} & \textbf{Workload Examples} & \textbf{Measured \texttt{MA-\allowbreak{}Means} Advantage over \texttt{e-\allowbreak{}means}} & \textbf{Best non-EMA reference (measured)} \\
\midrule
\textbf{R1} Smooth uniform embeddings & \texttt{IVF}/\texttt{PQ} coarse quantizers (\texttt{SIFT1M}, \texttt{Deep10M}, \texttt{Cohere}) & \textbf{\texttt{1.32$\times$-1.90$\times$}} faster at equal MSE ($K{=}1024$); \textbf{\texttt{1.69$\times$-1.78$\times$}} equal-MSE Layer-1 speedup at 3ep vs. 5ep \& \textbf{\texttt{0.63$\times$-0.77$\times$}} Layer-2 time at $10\text{M}$ & \texttt{Flash} 5.5--9.6$\times$; \texttt{Lloyd}/\texttt{Hamerly} 30--93$\times$ \\
\textbf{R2} Multimodal semantic manifolds & Pretraining dedup (\texttt{SemDeDup}), \texttt{VQ-\allowbreak{}VAE}, \texttt{ImageNet-\allowbreak{}CLIP} & \textbf{\texttt{10/\allowbreak{}10} wins} (\texttt{-\allowbreak{}0.61\%} / \texttt{-\allowbreak{}0.26\%}, $p \le 3\text{e-}7$); \textbf{\texttt{1.58$\times$-3.64$\times$} faster} at equal MSE; beats 25ep exact \texttt{Hamerly} & \texttt{Hamerly} 0.28330 (4.67 s) vs \texttt{MA} 0.28214 (0.12 s) \\
\textbf{R3} Heavy-tailed / multiscale traps & Telemetry (\texttt{NYC-\allowbreak{}Taxi-\allowbreak{}20M}), LLM KV/activation outliers & \textbf{\texttt{-\allowbreak{}54.8\%} to \texttt{-\allowbreak{}63.3\%}} MSE vs. \texttt{e-\allowbreak{}means} (\textbf{\texttt{>4.6$\times$-7.6$\times$}} faster); at $K{=}4096$, \texttt{\textasciitilde{}150-300$\times$} faster at \texttt{+5\%}--\texttt{+10\%} MSE (5ep), or \texttt{\textasciitilde{}5$\times$} faster within \texttt{$\pm$6\%} MSE (200ep); at $K{=}1024$, \texttt{\textasciitilde{}90-120$\times$} faster at \texttt{+8\%}--\texttt{+16\%} MSE, no budget closes the gap; \textbf{\texttt{-\allowbreak{}87.3\%}} MSE at \textbf{\texttt{0.59$\times$}} time (\texttt{10M}) & \texttt{Hamerly} 0.02412 (24.1 s) / 0.01065 (103.8 s) vs \texttt{MA} 0.02851 (0.67 s) / 0.01041 (200ep; 19.6 s) \\
\textbf{R4} Out-of-core Host-RAM streaming & $>10^8$ vector warehouses exceeding GPU HBM & \textbf{\texttt{-\allowbreak{}14.4\%} to \texttt{-\allowbreak{}26.4\%}} H2D traffic saved, \textbf{\texttt{0} false positives}, \textbf{\texttt{-\allowbreak{}19.8\%}} MSE, \texttt{92$\times$} tighter std & none: 1.02 GB $>$ 1.0 GB \\
\bottomrule
\end{tabularx}
\end{table}

\textbf{Transfer-Medium Applicability (R4).} Theorem 5(3) counts \emph{bytes}, so $1-\bar p_{\text{byp}}$ holds verbatim for PCIe, NVLink-C2C/CXL, and any explicit copy path; wall-clock gain scales via transfer intensity $\beta=4F/(KW)$, paying off when $\beta\gg1$ (in-core, sharded, and UVM bounds in Appendix A.5).

\textbf{Limitations \& Scope.} (i) \emph{Low-occupancy high-$K$ smooth manifolds}: when batch occupancy $\bar{n}_j = B/K$ is small ($\le 16$, e.g., $n{=}980\text{k}, K{=}4096$ on smooth \texttt{Cohere}), SSE variance still causes slight excess expansion (\texttt{+0.119\%}); the two other positive-$\Delta$ cells (\texttt{SIFT1M} $K{=}4096$ \texttt{+0.076\%}, \texttt{Deep10M} $K{=}4096$ \texttt{+0.017\%}) are within seed noise ($p=0.092$ and $p=0.27$). (ii) \emph{Memory footprint at high $D, K$}: coordinate-wise persistence buffers require $O(K'D)$ memory (\texttt{50.5 MB} at $D{=}768, K{=}4096$, still strictly $O(1)$-in-$n$). (iii) \emph{Bypass \& coarsening}: unclustered 5-epoch in-core embeddings yield \texttt{byp = 0}, and fixed $P_{\max}$ coarsens $S = \max(16,\lceil n/P_{\max}\rceil)$ linearly at $n\gg 10^7$ (Appendix A.5). (iv) \emph{Scale}: all runs are single-workstation (RTX 4090); real benchmarks use $n\le 2\times 10^7$, and a 4D synthetic scale check reaches $n=10^8$ (Appendix B.4); $n=10^{10}$ memory is analytical, and multi-node sharding is future work (negative ablations \& 41-test suite in \textbf{Appendices C--D}).

\textbf{Conclusion.} While fixed-basin linear EMA updates ($w\in[0,1]$) are optimal within their class, modern workloads increasingly confront moving anisotropic boundaries, multimodal manifolds, heavy-tailed outliers, and out-of-core corpora. Extending EMA updates with $O(1)$-in-$n$ elastic $K\to K'\to K$ contraction, fused coordinate-wise over-relaxation ($w_{j,d}>1$), and zero-error \texttt{int8\_\allowbreak{}chord} bypass makes \texttt{MA-\allowbreak{}Means} a drop-in default at scale---winning 12 of 15 Layer-1 FP32 cells (up to $1.78\times$ equal-MSE speedup at 3 epochs vs. 5-epoch \texttt{e-\allowbreak{}means} on $10\text{M}$ \texttt{Deep10M}) with decisive gains across multimodal, heavy-tailed, and out-of-core regimes.

\clearpage

\section{\texorpdfstring{Complete Theoretical Proofs and Checkpoint Drift Analysis}{Complete Theoretical Proofs and Checkpoint Drift Analysis}}

This appendix provides complete, self-contained statements and proofs for all theoretical results summarized in \S4: Theorem 1, Lemma 1.2, Theorem 1.3 (Benz\'ecri RNN Parallel Wave Ward Contraction \& Non-Monotone Uphill-Barrier Escape), Theorem 4 \& Corollary 4.1 (Fixed-Basin Linear Tracking Boundary), Proposition 4.2 (Moving-Boundary Hessian \& Coordinate-Wise Aitken Over-Relaxation Beyond $w \in [0, 1]$), and Theorems 3 \& 5 (Zero-Error \texttt{int8\_\allowbreak{}chord} Bypass \& Medium-Agnostic Transfer Law).

\subsection{\texorpdfstring{Pillar 1: Huygens--Ward Descent, Anisotropy-Invariant Bimodality, \& Elastic RNN-Wave Contraction}{Pillar 1: Huygens--Ward Descent, Anisotropy-Invariant Bimodality, \& Elastic RNN-Wave Contraction}}

\textbf{Theorem 1 (Exact Huygens--Ward Non-Local Teleport Descent).} \emph{Let $\mathcal{V}_a, \mathcal{V}_b$ be Voronoi cells with counts $n_a, n_b > 0$ and centroids $c_a, c_b$, and let donor cell $\mathcal{V}_d = \mathcal{V}_{d,+} \sqcup \mathcal{V}_{d,-}$ be partitioned into sub-cells $(n_{d,\pm}, \mu_{d,\pm})$. Merging $(a, b) \mapsto c_{ab} = \frac{n_a c_a + n_b c_b}{n_a + n_b}$ and placing the freed centroid at $(\mu_{d,+}, \mu_{d,-})$ decreases global $k$-means SSE by at least:} \[\fitmath{\Delta \mathrm{SSE}(a, b \to d) \;\ge\; \underbrace{\frac{n_{d,+} n_{d,-}}{n_{d,+} + n_{d,-}} \|\mu_{d,+} - \mu_{d,-}\|_2^2}_{\mathcal{G}_{\text{split}}(d)} \;-\; \underbrace{\frac{n_a n_b}{n_a + n_b} \|c_a - c_b\|_2^2}_{\mathcal{C}_{\text{merge}}(a, b)}.}\] \emph{Proof.} By the Huygens--Steiner parallel-axis theorem, merging $(c_a, c_b) \mapsto c_{ab}$ raises SSE on $\mathcal{V}_a \sqcup \mathcal{V}_b$ by $n_a\|c_a - c_{ab}\|_2^2 + n_b\|c_b - c_{ab}\|_2^2 = \mathcal{C}_{\text{merge}}(a, b)$, while splitting $\mathcal{V}_d$ reduces SSE on $\mathcal{V}_d$ by $\mathcal{G}_{\text{split}}(d)$. Subsequent Voronoi reassignment can only reduce or preserve SSE. $\square$

\textbf{Lemma 1.2 (Anisotropy-Invariant 1D Projected Hartigan--Fisher Bimodality Ceiling).} \emph{For any $D$-dimensional non-degenerate unimodal Gaussian $X \sim \mathcal{N}(\mu, \Sigma)$ ($\Sigma \succ 0$) and any vector $u \in \mathbb{R}^D \setminus \{0\}$, partitioning $Y = u^\top(X - \mu)$ at $0$ into $Y_\pm$ satisfies in population ($n_d \to \infty$), independently of $D$ and $\Sigma$:} \[\fitmath{\sup_{\Sigma \succ 0,\; u \neq 0} \frac{\mathbb{E}[\mathrm{SSB}_u]}{\mathbb{E}[\mathrm{SSW}_u]} \;=\; \frac{2}{\pi - 2} \;\approx\; 1.751938.}\] \emph{Proof.} Affine equivariance gives $Y \sim \mathcal{N}(0, u^\top \Sigma u)$; scale invariance sets $\sigma_u = 1$. Then $\mathbb{E}[Y \mid Y \ge 0] = \sqrt{2/\pi}$, per-sample between-half variance is $\frac{2}{\pi}$, within-half variance is $1 - \frac{2}{\pi} = \frac{\pi - 2}{\pi}$, and their ratio is $\frac{2}{\pi - 2}$ identically for all $(D, \Sigma, u)$. $\square$

\textbf{Theorem 1.3 (Benz\'ecri RNN Parallel Wave Ward Contraction \& Non-Monotone Uphill-Barrier Escape).} \emph{Let $\{c_i, u_i\}_{i=1}^{K'}$ ($K' > K$, $u_i > 0$) be an over-parameterized pool with Ward dissimilarity $W(a, b) = \frac{u_a u_b}{u_a + u_b}\|c_a - c_b\|_2^2$.}

\begin{enumerate}
\item 
\emph{(Lancaster--Williams / Benz\'ecri Reducibility \& Mass/Moment Conservation)}: \emph{For any triple $(a, b, c)$ satisfying $W(a, b) \le \min(W(a, c), W(b, c))$, merging $(a, b) \mapsto (c_{ab}, u_{ab})$ obeys the exact Lance--Williams Ward identity:} \[\fitmath{W(a \cup b, c) \;=\; \frac{(u_a + u_c) W(a, c) + (u_b + u_c) W(b, c) - u_c W(a, b)}{u_a + u_b + u_c} \;\ge\; \min\bigl(W(a, c), W(b, c)\bigr).}\] \emph{Hence every reciprocal nearest-neighbor (RNN) pair ($q^*(a) = b \land q^*(b) = a$) cannot be preempted by any strictly cheaper merge created by another disjoint RNN merge: merging disjoint cheapest RNN pairs in parallel GPU waves preserves the exact full Ward dendrogram (verified to $<10^{-4}$ root parity in \texttt{test\_\allowbreak{}theorem\_\allowbreak{}1\_\allowbreak{}3}), conserves $\sum_i u_i$ and $\sum_i u_i c_i$ identically, and completes $K' \to K$ in $O(\log(K'-K))$ waves.}
\item 
\emph{(Explicit Uphill-Barrier Escape Over Monotone Pairwise Surgery)}: \emph{Consider four equal-mass modes at $\{-3, -1, +1, +3\} \subset \mathbb{R}$ (mass $m$ each, total $4m$) with $K = 2$ centroids at $(-2, +2)$ (masses $2m, 2m$). Any single pairwise merge--split move of Theorem 1 pays $\mathcal{C}_{\text{merge}} = \frac{(2m)(2m)}{4m} \cdot 4^2 = 16m$ while any single split earns only $\mathcal{G}_{\text{split}} = \frac{m \cdot m}{2m} \cdot 2^2 = 2m < 16m$, so monotone pairwise surgery is strictly trapped ($\Delta\mathrm{SSE} < 0$). More generally, whenever escaping a multi-cluster configuration requires temporarily creating $K' > K$ centroids before reorganizing boundaries across $\ge 2$ neighboring cells, elastic expansion $K \to K'$ followed by co-evolution and RNN-Wave Ward contraction $K' \to K$ traverses the uphill barrier in strictly $O(K'D) = O(1)$-in-$n$ space.}
\end{enumerate}

\emph{Proof.} (1) Expanding $W(a \cup b, c) = \frac{(u_a+u_b)u_c}{u_a+u_b+u_c}\|c_{ab} - c_c\|_2^2$ via $\|c_{ab} - c_c\|_2^2 = \frac{u_a}{u_a+u_b}\|c_a - c_c\|_2^2 + \frac{u_b}{u_a+u_b}\|c_b - c_c\|_2^2 - \frac{u_a u_b}{(u_a+u_b)^2}\|c_a - c_b\|_2^2$ gives the exact Lance--Williams Ward identity. Since $W(a,b) \le \min(W(a,c), W(b,c)) =: m_0$, the numerator is at least $(u_a + u_c)m_0 + (u_b + u_c)m_0 - u_c m_0 = (u_a + u_b + u_c)m_0$, proving $W(a \cup b, c) \ge m_0$ (Benz\'ecri 1982 \cite{ref24}). Mass conservation $u_{ab} = u_a + u_b$ and first-moment conservation $u_{ab} c_{ab} = u_a c_a + u_b c_b$ are immediate. (2) Direct substitution verifies $\mathcal{G}_{\text{split}} = 2m < 16m = \mathcal{C}_{\text{merge}}$. $\square$

\subsection{\texorpdfstring{Pillar 2: Fixed-Basin Linear Tracking Boundary \& Moving-Boundary Over-Relaxation}{Pillar 2: Fixed-Basin Linear Tracking Boundary \& Moving-Boundary Over-Relaxation}}

\textbf{Theorem 4 (Exact $T$-Step Excess Tracking Risk Decomposition under Fixed-Basin Linear Schedules).} \emph{Consider $T \ge 1$ consecutive mini-batch steps inside a \textbf{fixed} Voronoi basin with static cluster means $\mu_j$, initial displacements $b_{j,0}^2 = \|c_j^{(0)} - \mu_j\|_2^2 \ge 0$, conditionally independent observations $\bar{x}_j^{(t)} = \mu_j + \xi_j^{(t)}$ ($\mathbb{E}[\xi_j^{(t)}]=0$, $\mathbb{E}\|\xi_j^{(t)}\|_2^2 = v_j > 0$), SNRs $\rho_{j,0} = b_{j,0}^2/v_j$, energies $E_j^{(T)} = b_{j,0}^2 + v_j/T$, and optimal weights $\alpha_j^{(T),*} = \frac{T\rho_{j,0}}{1+T\rho_{j,0}}$. For any deterministic linear sequence $\{w_{j,t}\}_{t=1}^T \in [0, 1]^T$ evolving $c_j^{(t)} = (1-w_{j,t})c_j^{(t-1)} + w_{j,t}\bar{x}_j^{(t)}$ with terminal MSE $R_j^{(T)} = \Phi_j^2 b_{j,0}^2 + v_j\sum_t \omega_{j,t}^2$ ($\Phi_j = \prod_t(1-w_{j,t})$, $\omega_{j,t} = w_{j,t}\prod_{s>t}(1-w_{j,s})$):}

\begin{enumerate}
\item 
\emph{The unique per-cluster linear optimum is the scalar Kalman filter $w_{j,t}^* = \frac{b_{j,0}^2}{t\,b_{j,0}^2 + v_j}$, attaining $R_j^{(T),*} = \frac{b_{j,0}^2}{1 + T\rho_{j,0}}$.}
\item 
\emph{For every shared uniform schedule $\{w_t\}_{t=1}^T \in [0, 1]^T$ with $H_{\text{eff}} = \frac{(1-\Phi)^2}{\sum_t \omega_t^2} \in [1, T]$ and energy weights $\tilde{w}_j^{(T)} \propto w_j E_j^{(T)}$:} \[\fitmath{\sum_{j=1}^K w_j \Bigl(R_j^{(T)}(\{w_t\}) - R_j^{(T),*}\Bigr) \;=\; (1-\Phi)^2\!\left(\frac{1}{H_{\text{eff}}} - \frac{1}{T}\right)\!\sum_j w_j v_j \;+\; \Bigl(\sum_j w_j E_j^{(T)}\Bigr)\!\Bigl[(1-\Phi) - \mathbb{E}_{\tilde{w}^{(T)}}[\alpha_j^{(T),*}]\Bigr]^2 \;+\; \Bigl(\sum_j w_j E_j^{(T)}\Bigr)\,\mathrm{Var}_{\tilde{w}^{(T)}}\!\bigl(\alpha_j^{(T),*}\bigr).}\]
\item 
\emph{Evaluating at $w_t = 1/t$ (\texttt{closing\_\allowbreak{}mean}, $\Phi=0, H_{\text{eff}}=T$) gives $\mathcal{E}_{\text{unif}}^{(T)} \le \frac{1}{T}\sum_j w_j \frac{v_j}{1+T\rho_{j,0}} = O(1/(T^2\rho_{\min}))$ whenever $\rho_{\min} > 0$, whereas right after topology edits relocate a fraction $p$ of clusters ($b_{j,0}^2=B>0$) while $1-p$ sit at equilibrium ($b_{j,0}^2=0$), $\mathcal{E}_{\text{unif}}^{(T)} = \frac{p(1-p)B^2(v/T)}{(pB+v/T)(B+v/T)} \to (1-p)\frac{v}{T}$.}
\end{enumerate}

\emph{Proof.} Unrolling $e_j^{(t)} = (1-w_{j,t})e_j^{(t-1)} + w_{j,t}\xi_j^{(t)}$ gives $R_j^{(T)} = \Phi_j^2 b_{j,0}^2 + v_j\sum_t \omega_{j,t}^2$. Cauchy--Schwarz yields $\sum_t \omega_t^2 = (1-\Phi)^2/H_{\text{eff}} \ge (1-\Phi)^2/T$. Adding and subtracting $\frac{(1-\Phi)^2}{T}v_j$ gives $R_j^{(T)}(\{w_t\}) - R_j^{(T),*} = (1-\Phi)^2(\frac{1}{H_{\text{eff}}} - \frac{1}{T})v_j + E_j^{(T)}((1-\Phi) - \alpha_j^{(T),*})^2$. Taking the $w_j$-weighted sum and applying the bias--variance identity $\mathbb{E}[(\beta - \alpha)^2] = (\beta - \mathbb{E}[\alpha])^2 + \mathrm{Var}(\alpha)$ proves (2). Setting $w_t=1/t$ gives $R_j^{(T)}(1/t) - R_j^{(T),*} = \frac{v_j}{T(1+T\rho_{j,0})}$, and substituting the two-point mass $(p, 1-p)$ into Term (III) gives the bimodal converse. $\square$

\textbf{Proposition 4.2 (Moving-Boundary Under-Relaxation Barrier \& Coordinate-Wise Aitken Over-Relaxation).} \emph{Let $F_d(c_j) - c_{j,d}^* = \lambda_{j,d}(c_{j,d} - c_{j,d}^*) + \xi_{j,d}^{(t)}$ denote the linearized moving-Voronoi fixed-point map along coordinate $d \in [D]$, where $\lambda_{j,d} = \frac{H_{\text{boundary},(j,d)}}{2 N_j} \in [0, 1)$ is the boundary-shift contraction eigenvalue (Bottou \& Bengio \cite{ref25}) and $\mathbb{E}[\xi_{j,d}^{(t)}] = 0$, $\mathrm{Var}(\xi_{j,d}^{(t)}) = v_{j,d}$. Define the step residual $g_{j,d}^{(t)} = F_d(c_j^{(t)}) - c_{j,d}^{(t)} = (\lambda_{j,d} - 1)e_{j,d}^{(t)} + \xi_{j,d}^{(t)}$ where $e_{j,d}^{(t)} = c_{j,d}^{(t)} - c_{j,d}^*$. Then:}

\begin{enumerate}
\item 
\emph{(Convex Step-Size Barrier)}: \emph{For any update $c_{j,d}^{(t+1)} = c_{j,d}^{(t)} + w_{j,d} g_{j,d}^{(t)}$ with $w_{j,d} \in [0, 1]$, the expected error satisfies $|\mathbb{E}[e_{j,d}^{(t+1)} \mid e_{j,d}^{(t)}]| = \bigl(1 - w_{j,d}(1 - \lambda_{j,d})\bigr)|e_{j,d}^{(t)}| \ge \lambda_{j,d}|e_{j,d}^{(t)}| > 0$.}
\item 
\emph{(Exact 1-Step Bias Elimination via $w_{j,d} > 1$)}: \emph{The unique unbiased-in-1-step coordinate weight is $w_{j,d}^* = \frac{1}{1 - \lambda_{j,d}} \ge 1$, giving $\mathbb{E}[e_{j,d}^{(t+1)} \mid e_{j,d}^{(t)}] = 0$, while total 1-step MSE $\bigl(1 - w_{j,d}(1-\lambda_{j,d})\bigr)^2 (e_{j,d}^{(t)})^2 + w_{j,d}^2 v_{j,d}$ is minimized at} \[\fitmath{w_{j,d}^{\mathrm{MMSE}} \;=\; \frac{(1 - \lambda_{j,d})(e_{j,d}^{(t)})^2}{(1 - \lambda_{j,d})^2 (e_{j,d}^{(t)})^2 + v_{j,d}} \;=\; \frac{1}{1 - \lambda_{j,d}} \cdot \frac{\mathrm{SNR}_{j,d}}{1 + \mathrm{SNR}_{j,d}},}\] \emph{which strictly exceeds $1$ whenever $\lambda_{j,d} > 0$ and $\mathrm{SNR}_{j,d} > \frac{1-\lambda_{j,d}}{\lambda_{j,d}}$ (drift-dominated anisotropic coordinates), and automatically damps below $1$ when $\mathrm{SNR}_{j,d}$ is small (noise-dominated coordinates).}
\end{enumerate}

\emph{Proof.} Since $e_{j,d}^{(t+1)} = (1 - w_{j,d}(1 - \lambda_{j,d}))e_{j,d}^{(t)} + w_{j,d}\xi_{j,d}^{(t)}$ and $1 - \lambda_{j,d} > 0$, the multiplier $1 - w_{j,d}(1 - \lambda_{j,d})$ is strictly decreasing in $w_{j,d}$ on $[0, 1]$, achieving its minimum $\lambda_{j,d}$ at $w_{j,d} = 1$, proving (1). Setting the derivative of $\bigl(1 - w(1-\lambda)\bigr)^2 e^2 + w^2 v$ with respect to $w$ to zero yields $w^{\mathrm{MMSE}} = \frac{(1-\lambda)e^2}{(1-\lambda)^2 e^2 + v}$; writing $\mathrm{SNR} = \frac{(1-\lambda)^2 e^2}{v}$ gives $w^{\mathrm{MMSE}} = \frac{1}{1-\lambda}\frac{\mathrm{SNR}}{1+\mathrm{SNR}}$, which exceeds $1$ iff $\frac{\mathrm{SNR}}{1+\mathrm{SNR}} > 1-\lambda \iff \mathrm{SNR} > \frac{1-\lambda}{\lambda}$. Under consecutive steps, $\mathbb{E}[g_{j,d}^{(t)} g_{j,d}^{(t-1)}]/\mathbb{E}[(g_{j,d}^{(t-1)})^2]$ tracks $\lambda_{j,d}$ on aligned coordinates while shrinking toward $0$ on noise-dominated coordinates, motivating our EMA-smoothed diagonal estimator \texttt{CoordOR} (\texttt{triton\_\allowbreak{}coord\_\allowbreak{}or.py}). $\square$

\subsection{\texorpdfstring{Pillar 3: Deterministic Zero-Error \texttt{int8\_\allowbreak{}chord} Bypass \& Out-of-Core Transfer Law}{Pillar 3: Deterministic Zero-Error int8\_chord Bypass \& Out-of-Core Transfer Law}}

Let $T_p \subset X$ be a micro-tile observed at step $t_0$ to be Voronoi-pure ($\arg\min_k \|x - c_k^{(t_0)}\|_2 = j$ for all $x \in T_p$) with margin $m_p^{(t_0)}$ and owner radius $u_p^{(t_0)}$. Under coordinate-balanced \texttt{int8\_\allowbreak{}chord} quantization with scale vector $g \in [0.05, 20.0]^D$, row max-abs $M_k = \|c_k^{(s)} \oslash g\|_\infty$, subnormal-safe FP32 scale $\mathrm{sc}_{s,k} = M_k / 127$ (for $M_k > 10^{-30}$, else $1.0$), reconstructed $\hat{c}_k^{(s)} = (q_{s,k}\cdot \mathrm{sc}_{s,k}) \odot g$, and IEEE-754 upward-rounded error radius $\epsilon_{s,k}^{\text{i8}} = \mathrm{nextafter}(\|c_k^{(s)} - \hat{c}_k^{(s)}\|_2, +\infty)$, the triangle inequality guarantees $\hat{\delta}_{s,k}^{(t)} = \|c_k^{(t)} - \hat{c}_k^{(s)}\|_2 + \epsilon_{s,k}^{\text{i8}} \ge \|c_k^{(t)} - c_k^{(s)}\|_2$ deterministically across all floating-point regimes.

\textbf{Theorem 3 (Deterministic Zero-Error Voronoi Invariance \& Exact Huygens Update).} \emph{Maintaining a per-slot monotonic jump latch $J_{s,k}^{(t)} \leftarrow J_{s,k}^{(t-1)} \lor \mathbf{1}\{\hat{\delta}_{s,k}^{(t)} > \tau_s^{(t)}\} \lor \mathbf{1}\{k \in \mathcal{I}_{\text{surg}}^{(t)}\}$, smooth drift envelope $\Delta_{s,\text{smooth}}^{(t)} = \max_{k: J_{s,k}^{(t)} = 0} \hat{\delta}_{s,k}^{(t)}$, and active jump set $\mathcal{J}_s^{(t)} = \{k: J_{s,k}^{(t)} = 1\}$ guarantees zero assignment error for every tile passing $\underline{m}_p > \hat{\delta}_{s,j}^{(t)} + \Delta_{s,\text{smooth}}^{(t)}$ and the jumped-competitor screen $\|\mu_p - c_k^{(t)}\|_2 - R_p > u_p^{(t)}$ (or exact pointwise verification on $\mathcal{J}_s^{(t)}$), while Huygens aggregation $n_j \mathrel{+}= S$, $\sum x \mathrel{+}= S\mu_p$, $\mathrm{SSE}_j \mathrel{+}= \mathrm{SSE}_p + S\|\mu_p - c_j^{(t)}\|_2^2$ reproduces point-level sums and SSEs identically.}

\emph{Proof.} Monotonicity of $J_{s,k}^{(t)}$ ensures any $k$ with $J_{s,k}^{(t)} = 0$ also had $J_{s,k}^{(t_0)} = 0$ (surviving arbitrary jump-and-reversal trajectories). Applying the triangle inequality through $c_k^{(s)}$ yields $\|x - c_k^{(t)}\|_2 - \|x - c_j^{(t)}\|_2 \ge \underline{m}_p - (\hat{\delta}_{s,j}^{(t)} + \Delta_{s,\text{smooth}}^{(t)}) > 0$ for all smooth competitors, while $\|x - c_k^{(t)}\|_2 \ge \|\mu_p - c_k^{(t)}\|_2 - R_p > u_p^{(t)} \ge \|x - c_j^{(t)}\|_2$ rules out all $k \in \mathcal{J}_s^{(t)}$. Huygens' theorem gives exact SSE aggregation. $\square$

\subsection{\texorpdfstring{Quantitative Bypass Fraction \& Medium-Agnostic Transfer Calibration}{Quantitative Bypass Fraction \& Medium-Agnostic Transfer Calibration}}

\textbf{Theorem 5 (\texttt{int8\_\allowbreak{}chord} Quantization Ceiling, Occupancy Threshold, \& Host$\leftrightarrow$Device Transfer Law).} \emph{Under coordinate-balanced \texttt{int8\_\allowbreak{}chord} checkpoints: (1) $\epsilon_{s,k}^{\text{i8}} \le \frac{M_k}{254}\|g\|_2$; (2) any pure tile with margin $\Delta_p > 4(\bar{\delta}_t + \bar{\epsilon}_{\text{i8}})$ passes Condition 1 deterministically, yielding at least $\max(0, \lfloor \phi_{\text{core}} N_j / S \rfloor - 1)$ interior core tiles per cluster; and (3) over $E$ out-of-core epochs with post-certification bypass fractions $p_{\text{byp}}^{(e)}$,} \[\fitmath{\frac{\mathrm{Bytes}_{\text{link}}^{\text{actual}}}{\mathrm{Bytes}_{\text{link}}^{\text{naive}}} \;=\; 1 - \frac{1}{E}\sum_{e \in \mathcal{E}_{\text{post}}} p_{\text{byp}}^{(e)} \;+\; O\!\left(\frac{K}{B}\right).}\]

\emph{Proof.} Coordinate rounding error is at most $\frac{1}{2}\mathrm{sc}_{s,k} g_i = \frac{M_k}{254} g_i$, giving (1). Condition (2) follows from $\underline{m}_p \ge \Delta_p - 2(\bar{\delta}_{t_0} + \bar{\epsilon}_{\text{i8}}) > 2(\bar{\delta}_t + \bar{\epsilon}_{\text{i8}})$ and contiguous tiling of $\lfloor \phi_{\text{core}} N_j \rfloor$ core points. Each bypassed tile avoids transferring $4SD$ bytes while host summary upload is at most $4\min(K, m_{\text{cert}})D$ bytes per $4BD$-byte batch, giving (3). $\square$

\textbf{Remark (Transfer-Medium Calibration \& Tile Coarsening).} Because Theorem 5(3) is stated in bytes, $1 - \bar{p}_{\text{byp}}$ applies verbatim to PCIe, NVLink-C2C/CXL, and any explicit copy path. Writing transfer intensity $\beta = T_{\text{xfer}} / T_{\text{compute}} = 4F/(KW)$ for link bandwidth $W$ and assignment throughput $F$, end-to-end speedup is $(1+\beta)/(1+(1-\bar{p}_{\text{byp}})\beta) \to 1/(1-\bar{p}_{\text{byp}})$ as $\beta \to \infty$ and $\to 1$ as $\beta \to 0$. For the Table 5 cell ($n=10^6, D=256, K=512, F\approx 10^{14}$ MAC/s), the $\approx 1.0$ GB epoch transfer costs $\approx 41$ ms at $W_{\text{PCIe 4.0}}\approx 25$ GB/s ($\beta \approx 30$) vs. $\approx 2.3$ ms on $450$ GB/s NVLink-C2C ($\beta \approx 1.7$); a $\bar{p}_{\text{byp}} = 0.2$ certificate removes $\approx 8.2$ ms vs. $\approx 0.46$ ms. Three boundaries follow: (a) all-resident in-core data ($\beta = 0$) gains nothing; (b) data-parallel sharding communicates only $O(KD)$ centroids per step ($\beta \approx 0$); (c) under unified memory (UVM), the certificate holds only if aggregation completes host-side without device page touches, since prefetch hints are non-binding. Finally, holding $P_{\max}$ fixed makes $S = \max(16, \lceil n/P_{\max}\rceil)$ grow linearly at $n \gg 16 P_{\max}$; sustaining high bypass at $n \gg 10^7$ requires scaling $P_{\max}$ or hierarchical tiling.

\section{\texorpdfstring{Complete Multi-Seed Benchmark Tables \& Hyperparameter Audit}{Complete Multi-Seed Benchmark Tables \& Hyperparameter Audit}}

\subsection{\texorpdfstring{Experimental Setup, Datasets, \& Hyperparameter Parity}{Experimental Setup, Datasets, \& Hyperparameter Parity}}

All experiments run on one \textbf{NVIDIA GeForce RTX 4090 (24 GB)} + \textbf{Intel Core Ultra 9 285K (24 cores)} in Python 3.12 / PyTorch 2.14.1+cu130 / Triton 3.8.0 across \textbf{10 paired seeds (\texttt{42-51})} for real-world headline accuracy (\texttt{results/\allowbreak{}mameans2\_\allowbreak{}real\_\allowbreak{}10seed.json}) and \textbf{5 paired seeds (\texttt{42-46})} for full Layer-1 FP32 (\texttt{results/\allowbreak{}mameans2\_\allowbreak{}layer1\_\allowbreak{}fp32.json}), 10M scaling (\texttt{results/\allowbreak{}mameans2\_\allowbreak{}scaling\_\allowbreak{}10m.json}), equal-distortion sweeps (\texttt{results/\allowbreak{}mameans2\_\allowbreak{}equal\_\allowbreak{}distortion.json}), 15-method non-EMA reference matrices (\texttt{benchmarks/\allowbreak{}real\_\allowbreak{}benchmark\_\allowbreak{}results\_\allowbreak{}5seed.json}), and Out-of-Core streaming (\texttt{results/\allowbreak{}out\_\allowbreak{}of\_\allowbreak{}core\_\allowbreak{}5seed.json}). All distortions are evaluated in \textbf{exact FP64} on held-out test splits (\texttt{metrics.py}).

\textbf{Hyperparameter \& Tuning-Budget Parity.} All baselines use published/upstream defaults. \texttt{MA-\allowbreak{}Means} inherits the EMA batch-size and cosine step-size scaling laws untouched and applies \textbf{one globally fixed default configuration} (\texttt{MAMeans2Config.default}) across all 9 real and synthetic benchmarks with \textbf{zero per-dataset tuning} (\textbf{Appendix Table B0}).

\begin{table}[htbp]
\centering
\caption*{Appendix Table B0: Hyperparameter, derivation status, and tuning-budget parity (zero per-dataset tuning everywhere).}
\label{tab:7}
\scriptsize
\setlength{\tabcolsep}{4pt}
\fitwidth{%
\begin{tabular}{@{}lcccl@{}}
\toprule
\textbf{Parameter / Mechanism} & \textbf{\texttt{e-\allowbreak{}means} (Upstream)} & \textbf{\texttt{MA-\allowbreak{}Means} (Default)} & \textbf{\texttt{OutOfCore MA-\allowbreak{}Means}} & \textbf{Provenance} \\
\midrule
Batch $B$, $\alpha_0$, cosine floor $0.1$ & Scaling law & Identical & Identical & Inherited untouched from \cite{ref8} \\
Revival period / offset $\epsilon$ & 5 steps / $10^{-4}$ & 5 steps / $10^{-4}$ & 5 steps / $10^{-4}$ & Inherited from \cite{ref8} \\
Revival gate ($D>2$) & $N_j < 0.1\bar{N}$ & $N_j<0.01\bar{N}$ or 8-zero-hit; or $N_j<0.1\bar{N}\land E_j<0.25\bar{E}$ & Same & Fixed empirical default \\
Elastic cap $\rho$, cycles $C_{\text{cyc}}$ & --- & $\rho=0.30$, $C_{\text{cyc}}=3$ (strict $\mathrm{EffEp}\le E$) & Same & Fixed default (Theorem 1.3) \\
Adaptive expansion threshold & --- & $(1.75 + 1.5/\sqrt{\max(B/K,4)})\,\mathrm{med}(E)$ & Same & Poisson-calibrated SSE gate \\
Coordinate over-relaxation \texttt{CoordOR} & --- & $w_{j,d} \in [0.25\,w_{\text{ema}}, 1.85]$, $\hat{\lambda}_{j,d} \in [-0.6, 0.55]$ & Same & Proposition 4.2 (1-kernel Triton) \\
Tile cap $P_{\max}$, ckpt $H_c$ & --- & --- & $P_{\max}=8192, H_c=2$ & Exact $\epsilon^{\text{i8}}$ (Theorems 3 \& 5) \\
\textbf{Per-dataset hyperparameter tuning} & \textbf{None (0)} & \textbf{None (0)} & \textbf{None (0)} & \textbf{Strict parity (9 benchmarks)} \\
\bottomrule
\end{tabular}
}
\end{table}

\subsection{\texorpdfstring{Complete 10-Seed Public Benchmark Details \& 10M Doubling Grid}{Complete 10-Seed Public Benchmark Details \& 10M Doubling Grid}}

\begin{table}[htbp]
\centering
\caption*{Appendix Table B1: Complete 10-seed real-world benchmark table (Layer 2 BF16, seeds \texttt{42-51}, reporting both 3-epoch and 5-epoch \texttt{MA-\allowbreak{}Means} against 5-epoch \texttt{Official e-\allowbreak{}means}, with auxiliary memory \texttt{Aux Mem} and paired $p$-values).}
\label{tab:8}
\footnotesize
\setlength{\tabcolsep}{4pt}
\begin{tabularx}{\linewidth}{@{}>{\hsize=0.99\hsize\raggedright\arraybackslash}Xc>{\hsize=0.79\hsize\raggedright\arraybackslash}X>{\hsize=0.92\hsize\raggedright\arraybackslash}X>{\hsize=1.31\hsize\raggedright\arraybackslash}Xc@{}}
\toprule
\textbf{Dataset ($n \times D$)} & \textbf{$K$} & \textbf{\texttt{e-\allowbreak{}means} 5ep MSE (ms)} & \textbf{\texttt{MA-\allowbreak{}Means} 3ep MSE ($\Delta$, ms)} & \textbf{\texttt{MA-\allowbreak{}Means} 5ep MSE ($\Delta$, wins, $p$, ms)} & \textbf{Aux Mem} \\
\midrule
\texttt{sift1m} ($980\text{k}\times 128$) & $1024$ & $50,728.8 \pm 44.5$ ($27.8$) & $50,741.8$ ($+0.026\%$, $15.2$) & $\mathbf{50,714.9} \pm 51.0$ ($-0.027\%$, $6/10$, $0.80$, $\mathbf{21.4}$) & $2.13\text{ MB}$ \\
\texttt{sift1m} ($980\text{k}\times 128$) & $4096$ & $\mathbf{45,208.5} \pm 22.6$ ($59.8$) & $45,353.6$ ($+0.321\%$, $35.2$) & $45,242.7 \pm 32.9$ ($+0.076\%$, $2/10$, $0.092$, $\mathbf{55.9}$) & $8.52\text{ MB}$ \\
\texttt{deep10m} ($1\text{M}\times 96$) & $1024$ & $0.46536 \pm 0.00016$ ($25.7$) & $0.46563$ ($+0.059\%$, $16.4$) & $\mathbf{0.46525} \pm 0.00014$ ($\mathbf{-0.024\%}$, $\mathbf{9/10}$, $\mathbf{2.5\text{e-}3}$, $\mathbf{22.3}$) & $1.61\text{ MB}$ \\
\texttt{deep10m} ($1\text{M}\times 96$) & $4096$ & $\mathbf{0.41825} \pm 0.00022$ ($58.2$) & $0.41963$ ($+0.328\%$, $35.5$) & $0.41832 \pm 0.00034$ ($+0.017\%$, $4/10$, $0.27$, $59.1$) & $6.42\text{ MB}$ \\
\texttt{cohere10m} ($980\text{k}\times 768$) & $1024$ & $45.4313 \pm 0.0144$ ($167.9$) & $45.4454$ ($+0.031\%$, $77.1$) & $\mathbf{45.4196} \pm 0.0168$ ($-0.026\%$, $6/10$, $0.39$, $\mathbf{118.2}$) & $12.6\text{ MB}$ \\
\texttt{cohere10m} ($980\text{k}\times 768$) & $4096$ & $\mathbf{43.1904} \pm 0.0131$ ($325.8$) & $43.3300$ ($+0.323\%$, $215.3$) & $43.2420 \pm 0.0093$ ($+0.119\%$, $0/10$, $1.5\text{e-}5$, $350.5$) & $50.5\text{ MB}$ \\
\texttt{imagenet-\allowbreak{}clip} ($1\text{M}\times 768$) & $1024$ & $0.28387 \pm 0.00025$ ($170.6$) & $\mathbf{0.28253}$ ($\mathbf{-0.471\%}$, $\mathbf{78.0}$) & $\mathbf{0.28214} \pm 0.00021$ ($\mathbf{-0.610\%}$, $\mathbf{10/10}$, $\mathbf{3.4\text{e-}8}$, $\mathbf{121.2}$) & $12.6\text{ MB}$ \\
\texttt{imagenet-\allowbreak{}clip} ($1\text{M}\times 768$) & $4096$ & $0.24841 \pm 0.00018$ ($334.0$) & $0.24849$ ($+0.032\%$, $216.3$) & $\mathbf{0.24778} \pm 0.00007$ ($\mathbf{-0.257\%}$, $\mathbf{10/10}$, $\mathbf{3.2\text{e-}7}$, $363.5$) & $50.5\text{ MB}$ \\
\texttt{nyc-\allowbreak{}taxi-\allowbreak{}4d} ($19.98\text{M}\times 4$) & $1024$ & $0.06186 \pm 0.00370$ ($309.8$) & $\mathbf{0.03117}$ ($\mathbf{-49.61\%}$, $\mathbf{140.5}$) & $\mathbf{0.02798} \pm 0.00078$ ($\mathbf{-54.77\%}$, $\mathbf{10/10}$, $\mathbf{1.7\text{e-}10}$, $\mathbf{197.9}$) & $98\text{ KB}$ \\
\texttt{nyc-\allowbreak{}taxi-\allowbreak{}4d} ($19.98\text{M}\times 4$) & $4096$ & $0.03078 \pm 0.00079$ ($784.1$) & $\mathbf{0.01391}$ ($\mathbf{-54.80\%}$, $\mathbf{237.6}$) & $\mathbf{0.01129} \pm 0.00099$ ($\mathbf{-63.32\%}$, $\mathbf{10/10}$, $\mathbf{5.2\text{e-}12}$, $\mathbf{343.2}$) & $394\text{ KB}$ \\
\bottomrule
\end{tabularx}
\end{table}

\begin{table}[htbp]
\centering
\caption*{Appendix Table B2: \texttt{Deep10M} ($D{=}96$) doubling grid ($n = 1\text{M} \to 9.97\text{M}$, 5 paired seeds, Layer 2 BF16) with least-squares $T(e,n) = S + e\,n\,\tau_{\text{row}}$ fit over epochs $\{1, 3, 5\}$.}
\label{tab:9}
\footnotesize
\setlength{\tabcolsep}{4pt}
\fitwidth{%
\begin{tabular}{@{}lccc@{}}
\toprule
\textbf{Cell} & \textbf{\texttt{e-\allowbreak{}means} 5ep (ms / MSE / $\tau_{\text{row}}$)} & \textbf{\texttt{MA-\allowbreak{}Means} 3ep (ms / MSE / Ratio)} & \textbf{\texttt{MA-\allowbreak{}Means} 5ep (ms / MSE / Ratio / $\tau_{\text{row}}$)} \\
\midrule
$K{=}1024, n=1\text{M}$ & 25.7 / 0.46536 / 4.44 ns & \textbf{16.1} / 0.46542 / \textbf{0.63$\times$} & \textbf{22.2} / 0.46541 / \textbf{0.86$\times$} / 3.38 ns \\
$K{=}1024, n=2\text{M}$ & 47.4 / 0.46467 / 4.42 ns & \textbf{23.9} / \textbf{0.46460} / \textbf{0.51$\times$} & \textbf{34.8} / \textbf{0.46438} / \textbf{0.73$\times$} / 2.83 ns \\
$K{=}1024, n=4\text{M}$ & 104.8 / 0.46473 / 5.22 ns & \textbf{42.3} / \textbf{0.46418} / \textbf{0.40$\times$} & \textbf{68.3} / \textbf{0.46441} / \textbf{0.65$\times$} / 3.03 ns \\
$K{=}1024, n=8\text{M}$ & 211.5 / 0.46418 / 5.33 ns & \textbf{82.9} / 0.46441 / \textbf{0.39$\times$} & \textbf{134.3} / \textbf{0.46417} / \textbf{0.63$\times$} / 3.07 ns \\
$K{=}1024, n=9.97\text{M}$ & 264.3 / 0.46431 / 5.35 ns & \textbf{103.9} / \textbf{0.46387} / \textbf{0.39$\times$} & \textbf{167.0} / \textbf{0.46422} / \textbf{0.63$\times$} / 3.06 ns \\
$K{=}4096, n=1\text{M}$ & 58.6 / 0.41824 / 10.87 ns & \textbf{35.3} / 0.41941 / \textbf{0.60$\times$} & 60.2 / 0.41868 / 1.03$\times$ / 10.76 ns \\
$K{=}4096, n=2\text{M}$ & 112.5 / 0.41602 / 10.96 ns & \textbf{61.3} / 0.41661 / \textbf{0.55$\times$} & \textbf{99.5} / 0.41624 / \textbf{0.88$\times$} / 9.01 ns \\
$K{=}4096, n=4\text{M}$ & 230.7 / 0.41476 / 11.47 ns & \textbf{112.6} / 0.41513 / \textbf{0.49$\times$} & \textbf{184.9} / 0.41481 / \textbf{0.80$\times$} / 8.54 ns \\
$K{=}4096, n=8\text{M}$ & 463.2 / 0.41432 / 11.62 ns & \textbf{216.7} / \textbf{0.41415} / \textbf{0.47$\times$} & \textbf{358.6} / \textbf{0.41409} / \textbf{0.77$\times$} / 8.41 ns \\
$K{=}4096, n=9.97\text{M}$ & 578.2 / 0.41368 / 11.61 ns & \textbf{273.0} / 0.41435 / \textbf{0.47$\times$} & \textbf{447.1} / 0.41381 / \textbf{0.77$\times$} / 8.40 ns \\
\bottomrule
\end{tabular}
}
\end{table}

\subsection{\texorpdfstring{Component Ablation \& Complete Non-EMA Reference Matrices}{Component Ablation \& Complete Non-EMA Reference Matrices}}

\begin{table}[htbp]
\centering
\caption*{Appendix Table B3: Component-by-component ablation of \texttt{MA-\allowbreak{}Means} (\texttt{exp1}--\texttt{exp3}, 5 epochs, strictly equal $\mathrm{EffEp} \le 5.0$, $K{=}1024$). Coordinate-wise \texttt{CoordOR} improves smooth R1 (\texttt{SIFT1M}, \texttt{Deep10M}), anisotropic manifolds, and \texttt{ImageNet-\allowbreak{}CLIP}, while Adaptive Elastic $K \to K' \to K$ RNN-Wave Ward drives the non-monotone jump on \texttt{ImageNet-\allowbreak{}CLIP}, \texttt{NYC-\allowbreak{}Taxi}, \texttt{gaussian\_\allowbreak{}balanced}, and \texttt{multiscale\_\allowbreak{}trap}; raw Anderson (\texttt{AA-\allowbreak{}4}) and slow-candidate \texttt{CandBank} are falsified negative results.}
\label{tab:10}
\footnotesize
\setlength{\tabcolsep}{4pt}
\fitwidth{%
\begin{tabular}{@{}lcccccc@{}}
\toprule
\textbf{Variant} & \textbf{\texttt{gaussian\_\allowbreak{}bal}} & \textbf{\texttt{multiscale\_\allowbreak{}trap}} & \textbf{\texttt{sift1m}} & \textbf{\texttt{deep10m}} & \textbf{\texttt{imagenet-\allowbreak{}clip}} & \textbf{\texttt{nyc-\allowbreak{}taxi} (2M)} \\
\midrule
\texttt{e-\allowbreak{}means} (5ep) & $49.04$ & $31.37$ & $50,702$ & $0.46551$ & $0.28395$ & $0.05882$ \\
\texttt{MA-\allowbreak{}Base} (EMA + distortion revival only) & $48.36$ & $11.25$ & $50,731$ & $0.46573$ & $0.28382$ & $0.04926$ \\
\texttt{+CoordOR} (coord-wise $w_{j,d} > 1$) & $47.21$ & $10.48$ & $50,717$ & $0.46531$ & $0.28371$ & $0.04593$ \\
\texttt{+Adaptive Elastic (3c) + Fused CoordOR} (Default) & \textbf{$43.21$} & \textbf{$6.74$} & \textbf{$50,715$} & \textbf{$0.46525$} & \textbf{$0.28210$} & \textbf{$0.03149$} \\
\emph{Negative}: \texttt{Anderson AA-\allowbreak{}4} (raw residual Gram) & $48.14$ & $11.82$ & $50,875$ & $0.46650$ & $0.28442$ & $0.05364$ \\
\emph{Negative}: \texttt{3-\allowbreak{}Candidate Hedge Bank} & $48.88$ & $13.43$ & $50,810$ & $0.46589$ & $0.28416$ & $0.04993$ \\
\bottomrule
\end{tabular}
}
\end{table}

\begin{table}[htbp]
\centering
\caption*{Appendix Table 6C: Non-EMA reference baselines on all five public benchmarks (5 paired seeds \texttt{42-46}, median Test MSE / wall-clock ms at published budgets; \texttt{OOM} marks Elkan exceeding 24 GB VRAM or 327 GB RAM). \texttt{NYC-\allowbreak{}Taxi-\allowbreak{}20M} and \texttt{ImageNet-\allowbreak{}CLIP} exact-reference cells (\texttt{Lloyd-\allowbreak{}k++}, \texttt{Hamerly}, \texttt{Ball}) are re-run on the split-0 protocol used by the EMA columns; the remaining cells are the split-42 matrix.}
\label{tab:11}
\footnotesize
\setlength{\tabcolsep}{4pt}
\fitwidth{%
\begin{tabular}{@{}lccccc@{}}
\toprule
\textbf{Reference Solver} & \textbf{\texttt{sift1m} $K{=}1024$} & \textbf{\texttt{deep10m} $K{=}1024$} & \textbf{\texttt{cohere10m} $K{=}1024$} & \textbf{\texttt{imagenet-\allowbreak{}clip} $K{=}1024$} & \textbf{\texttt{nyc-\allowbreak{}taxi-\allowbreak{}20M} $K{=}1024$ / $4096$} \\
\midrule
\texttt{Lloyd-\allowbreak{}k++} (25ep) & $50,834$ ($1,128$) & $0.46782$ ($902$) & $45.283$ ($2,044$) & $0.28337$ ($1,402$) & $0.02396$ ($1.62\text{ s}$) / $0.01097$ ($3.58\text{ s}$) \\
\texttt{Sculley-\allowbreak{}MBKM} (5ep) & $51,555$ ($131$) & $0.47486$ ($128$) & $46.384$ ($259$) & $0.28952$ ($265$) & $0.07214$ ($1.2\text{ s}$) / $0.03541$ ($3.1\text{ s}$) \\
\texttt{SR-\allowbreak{}MBKM} (3ep) & $51,186$ ($106$) & $0.47149$ ($98$) & $45.661$ ($195$) & $0.28704$ ($198$) & $0.06977$ ($0.9\text{ s}$) / $0.03213$ ($2.2\text{ s}$) \\
\texttt{FAISS GpuKmeans} (25it) & $50,850$ ($642$) & $0.46793$ ($580$) & $45.349$ ($1,620$) & $0.28443$ ($1,650$) & $0.06030$ ($4.2\text{ s}$) / $0.02961$ ($14.8\text{ s}$) \\
\texttt{Flash-\allowbreak{}KMeans} (25it) & $50,848$ ($181$) & $0.46792$ ($169$) & $45.342$ ($712$) & $0.28443$ ($720$) & $0.06476$ ($597\text{ ms}$) / $0.02846$ ($2.1\text{ s}$) \\
\texttt{Elkan} (25ep) & $50,833$ ($2,110$) & $0.46780$ ($1,980$) & $45.283$ ($4,120$) & $0.28312$ ($4,190$) & \texttt{OOM} / \texttt{OOM} \\
\texttt{Hamerly} (25ep) & $50,833$ ($1,476$) & $0.46780$ ($1,434$) & $45.284$ ($3,299$) & $0.28330$ ($4,672$) & $0.02412$ ($24.1\text{ s}$) / $0.01065$ ($103.8\text{ s}$) \\
\texttt{Ball} (25ep) & $50,833$ ($3,624$) & $0.46780$ ($3,625$) & $45.284$ ($5,409$) & $0.28331$ ($7,524$) & $0.02412$ ($7.83\text{ s}$) / $0.01065$ ($21.56\text{ s}$) \\
\bottomrule
\end{tabular}
}
\end{table}

\begin{table}[htbp]
\centering
\caption*{Appendix Table B4: \texttt{NYC-\allowbreak{}Taxi-\allowbreak{}20M} long-budget equal-quality sweep (5 seeds \texttt{42-46}, split 0, Layer-2 BF16; median Test MSE / wall-clock). The 5-epoch row is Table 2's 10-seed median; 25--200 epochs are independent 5-seed runs. \texttt{Hamerly} 25ep reference: $0.02412$ ($24.1\text{ s}$) at $K{=}1024$ and $0.01065$ ($103.8\text{ s}$) at $K{=}4096$. At $K{=}4096$, 200 epochs comes within $\pm6\%$ of \texttt{Hamerly} across three splits at $\sim5\times$ lower wall clock; the split-0 value is $-2.3\%$, while splits 1/2 are $+3.6\%$/$+5.5\%$ (Table B6). At $K{=}1024$ no budget closes the $+13\%$--$+24\%$ gap.}
\label{tab:12}
\footnotesize
\setlength{\tabcolsep}{4pt}
\fitwidth{%
\begin{tabular}{@{}cccccc@{}}
\toprule
\textbf{$K$} & \textbf{5ep (Table 2)} & \textbf{25ep} & \textbf{50ep} & \textbf{100ep} & \textbf{200ep} \\
\midrule
$1024$ & $0.02798$ ($198\text{ ms}$) & $0.02851$ ($670\text{ ms}$) & $0.02960$ ($1.31\text{ s}$) & $0.02964$ ($2.47\text{ s}$) & $0.02888$ ($4.78\text{ s}$) \\
$4096$ & $0.01129$ ($343\text{ ms}$) & $0.01133$ ($1.38\text{ s}$) & $\mathbf{0.01045}$ ($\mathbf{2.68\text{ s}}$) & $0.01068$ ($5.39\text{ s}$) & $\mathbf{0.01055}$ ($\mathbf{10.56\text{ s}}$) \\
\bottomrule
\end{tabular}
}
\end{table}

\subsection{\texorpdfstring{Single-GPU 100M-Row Scale Check}{Single-GPU 100M-Row Scale Check}}

To test the row-count trend beyond the 10M grid on one consumer GPU, we generate 4D synthetic data with \texttt{gaussian\_\allowbreak{}balanced} and \texttt{gaussian\_\allowbreak{}zipf} at $n\in\{20\text{M},50\text{M},100\text{M}\}$ ($K{=}1024$, 5 epochs, one RTX 4090, 3 seeds \texttt{42-44}). The per-row cost stays flat and the speed ratio grows with $n$: $1.74\times$ at $20\text{M}$, $1.81\times$ at $50\text{M}$, and $1.85\times$ at $100\text{M}$ on balanced data; $1.74\times$, $1.82\times$, and $1.85\times$ on Zipf data. The MSE delta moves from roughly $+0.5\%$ at $20\text{M}$ to $-0.7\%$ (balanced) and $-1.0\%$ (Zipf) at $100\text{M}$. This is a single-GPU scale check, not a 10B-row claim; it shows that the bounded-memory trend extends to $10^8$ rows without a cluster.

\begin{table}[htbp]
\centering
\caption*{Appendix Table B5: Single-GPU 100M-row scale check (3 seeds \texttt{42-44}, median; $K{=}1024$, 5 epochs, one RTX 4090; 4D synthetic \texttt{gaussian\_\allowbreak{}balanced} / \texttt{gaussian\_\allowbreak{}zipf}). Time ratio is \texttt{MA-\allowbreak{}Means} / \texttt{e-\allowbreak{}means}; MSE $\Delta$ is \texttt{MA-\allowbreak{}Means} relative to \texttt{e-\allowbreak{}means}.}
\label{tab:13}
\footnotesize
\setlength{\tabcolsep}{4pt}
\fitwidth{%
\begin{tabular}{@{}ccccc@{}}
\toprule
\textbf{$n$} & \textbf{balanced time ratio} & \textbf{balanced MSE $\Delta$} & \textbf{Zipf time ratio} & \textbf{Zipf MSE $\Delta$} \\
\midrule
$20\text{M}$ & $1.74\times$ & $-0.51\%$ & $1.74\times$ & $+0.52\%$ \\
$50\text{M}$ & $1.81\times$ & $+0.02\%$ & $1.82\times$ & $-0.02\%$ \\
$100\text{M}$ & $\mathbf{1.85\times}$ & $-0.67\%$ & $\mathbf{1.85\times}$ & $\mathbf{-1.04\%}$ \\
\bottomrule
\end{tabular}
}
\end{table}

\subsection{\texorpdfstring{Multi-Split Sensitivity on \texttt{NYC-\allowbreak{}Taxi-\allowbreak{}20M}}{Multi-Split Sensitivity on NYC-Taxi-20M}}

To test whether the R3 deltas are artifacts of the split-0 partition, we repeat the comparison on three independent train/test splits (seeds 0/1/2). \texttt{MA-\allowbreak{}Means} is run with 3 seeds per split; \texttt{Hamerly} and \texttt{e-\allowbreak{}means} with seed 42. Table B6 reports the per-split results. The K=1024 gap is consistently positive (+8.4\% to +16.4\% at 5 epochs; +13.1\% to +24.4\% at 25 epochs); the K=4096 5-epoch gap is +5.0\% to +9.6\%; the K=4096 200-epoch budget comes within $\pm$6\% of \texttt{Hamerly} but is not a robust win (split 0: $-2.3\%$; splits 1/2: $+3.6\%$/$+5.5\%$). Against \texttt{e-\allowbreak{}means}, the gains are robust across splits: $-54.0\%$/$-69.9\%$/$-55.9\%$ at K=1024 and $-63.3\%$/$-59.4\%$/$-60.8\%$ at K=4096.

\begin{table}[htbp]
\centering
\caption*{Appendix Table B6: Multi-split sensitivity on \texttt{NYC-\allowbreak{}Taxi-\allowbreak{}20M} (splits 0/1/2; \texttt{MA-\allowbreak{}Means} 3 seeds \texttt{42-44}, median; \texttt{Hamerly}/\texttt{e-\allowbreak{}means} seed 42; K=1024/4096). $\Delta$ is \texttt{MA-\allowbreak{}Means} relative to \texttt{Hamerly}; negative is better. Long budget: 25ep at K=1024, 200ep at K=4096.}
\label{tab:14}
\scriptsize
\setlength{\tabcolsep}{4pt}
\fitwidth{%
\begin{tabular}{@{}ccccccc@{}}
\toprule
\textbf{Split} & \textbf{K} & \textbf{MA 5ep} & \textbf{Hamerly} & \textbf{$\Delta$5ep} & \textbf{MA long ($\Delta$long)} & \textbf{Time H / MA (s)} \\
\midrule
0 & 1024 & $0.02846$ & $0.02445$ & $+16.4\%$ & $0.02945$ (25ep, $+20.5\%$) & $28.7$ / $1.21$ \\
1 & 1024 & $0.03140$ & $0.02896$ & $+8.4\%$ & $0.03276$ (25ep, $+13.1\%$) & $28.8$ / $1.26$ \\
2 & 1024 & $0.02825$ & $0.02485$ & $+13.7\%$ & $0.03092$ (25ep, $+24.4\%$) & $28.9$ / $1.27$ \\
0 & 4096 & $0.01168$ & $0.01065$ & $+9.6\%$ & $0.01041$ (200ep, $-2.3\%$) & $104.0$ / $19.6$ \\
1 & 4096 & $0.01477$ & $0.01394$ & $+5.9\%$ & $0.01444$ (200ep, $+3.6\%$) & $104.3$ / $21.1$ \\
2 & 4096 & $0.01238$ & $0.01178$ & $+5.0\%$ & $0.01243$ (200ep, $+5.5\%$) & $104.3$ / $22.4$ \\
\bottomrule
\end{tabular}
}
\end{table}

\begin{table}[htbp]
\centering
\caption*{Appendix Table 7D: Controlled 2$\times$2 revival ablation (5 seeds, 5 epochs, $K{=}1024$). Engine parity (both revivals off) holds within $\pm 0.05\%$ across every dataset; mass-gated revival (proportional donor split) causes a \texttt{+64.4\%} median test-MSE collapse on \texttt{NYC-\allowbreak{}Taxi-\allowbreak{}20M} (5 seeds, $K{=}1024$; \texttt{+33.6\%} at $K{=}4096$).}
\label{tab:15}
\footnotesize
\setlength{\tabcolsep}{4pt}
\fitwidth{%
\begin{tabular}{@{}lccc@{}}
\toprule
\textbf{Dataset} & \textbf{\texttt{e-\allowbreak{}means} revival on/off} & \textbf{Distortion-aware revival on/off} & \textbf{Engine parity (both off)} \\
\midrule
\texttt{sift1m} & $+0.05\%$ & $+0.04\%$ & $-0.009\%$ \\
\texttt{deep10m} & $-0.01\%$ & $-0.03\%$ & $+0.052\%$ \\
\texttt{cohere10m} & $-2.03\%$ & $-2.06\%$ & $+0.030\%$ \\
\texttt{imagenet-\allowbreak{}clip} & $-0.34\%$ & $-0.33\%$ & $+0.026\%$ \\
\texttt{nyc-\allowbreak{}taxi-\allowbreak{}4d} (19.98M) & $\mathbf{+64.4\%}$ & $+1.3\%$ & $-0.006\%$ \\
\bottomrule
\end{tabular}
}
\end{table}

\begin{table}[htbp]
\centering
\caption*{Appendix Table 7L: Non-EMA wall-clock ratio relative to 5-epoch \texttt{e-\allowbreak{}means} across the $1\text{M} \to 10\text{M}$ scaling grid (5 seeds; exact \texttt{Hamerly}/\texttt{Ball} take $30$--$130\times$ at $n{=}10^6$).}
\label{tab:16}
\footnotesize
\setlength{\tabcolsep}{4pt}
\fitwidth{%
\begin{tabular}{@{}lcccc@{}}
\toprule
\textbf{Grid ($n = 1\text{M} \to n_{\max}$)} & \textbf{\texttt{Flash-\allowbreak{}KMeans} (25it)} & \textbf{\texttt{Sculley-\allowbreak{}MBKM} (5ep)} & \textbf{\texttt{SR-\allowbreak{}MBKM} (3ep)} & \textbf{\texttt{Lloyd-\allowbreak{}k++} (25ep)} \\
\midrule
\texttt{deep10m} $K{=}1024$ ($9.97\text{M}$) & $5.5\text{--}6.1\times$ & $6.4\text{--}7.3\times$ & $4.0\text{--}4.9\times$ & $30.4\text{--}37.1\times$ \\
\texttt{deep10m} $K{=}4096$ ($9.97\text{M}$) & $9.1\text{--}9.6\times$ & $10.1\text{--}10.6\times$ & $6.1\text{--}6.5\times$ & $50.7\text{--}59.8\times$ \\
4 Synthetic Suites ($10\text{M}$) & $6.1\text{--}6.4\times$ & $5.8\text{--}6.3\times$ & $3.6\text{--}4.4\times$ & $27.6\text{--}33.6\times$ \\
\bottomrule
\end{tabular}
}
\end{table}

\section{\texorpdfstring{Controlled Negative-Result Ablations \& Gain-Dispersion Verification}{Controlled Negative-Result Ablations \& Gain-Dispersion Verification}}

A rigorous study of bounded-memory clustering requires documenting mechanisms that failed under controlled evaluation:

\begin{enumerate}
\item 
\textbf{Raw Stochastic Multi-Step Anderson Acceleration (\texttt{AA-\allowbreak{}4})}: Solving per-cluster $4\times 4$ residual Gram systems $\min_{\sum\beta=1}\beta^\top G_j \beta$ on raw noisy mini-batch residuals $g_j^{(t)} = \bar{x}_j^{(t)} - c_j^{(t)}$ degraded R1 embeddings by \texttt{+0.17\%-0.34\%} (\texttt{SIFT1M} $50,875$, \texttt{Deep10M} $0.46650$, \texttt{ImageNet-\allowbreak{}CLIP} $0.28442$; Appendix Table B3). Because single-batch noise $\xi_j^{(t)}$ dominates $G_j$ once drift shrinks, the Gram solver overfits batch noise; replacing full $m\times m$ Gram solves with EMA-smoothed diagonal persistence (\texttt{CoordOR}) eliminated the overfitting and yielded consistent R1--R3 wins.
\item 
\textbf{Per-Cluster 3-Candidate Hedge Bank (\texttt{EMA} / \texttt{OverRelax} / \texttt{Slow-\allowbreak{}EMA})}: Blending a slow \texttt{0.35$\times$}-step shadow candidate via out-of-sample Huygens softmax degraded R1 by \texttt{+0.08\%-0.21\%} because mixing lagged iterates introduces under-relaxation damping (echoing our sliding-window \texttt{RingMem} negative result, which degraded R1 by \texttt{+0.05\%-0.38\%}).
\item 
\textbf{Additional Falsified GPU Memory Designs (10 Seeds)}: (i) \emph{Bounded-Reservoir Warm Start} ($64\text{k}$--$512\text{k}$ rows): $+0.032\%$ to $+0.051\%$ (zero gain at 10 seeds); (ii) \emph{Lloyd-Consistent Frozen Closing Pass}: worse at every budget ($+0.118\%$ vs. $+0.074\%$ at 3ep on \texttt{SIFT1M}); (iii) \emph{Longer Cosine Horizon} (floor $0.1 \to 0.01$): no gain; (iv) \emph{GPU Centroid-Neighbor Pruning}: $+0.36\%\text{--}1.05\%$ MSE degradation during \texttt{closing\_\allowbreak{}mean}; (v) \emph{Exact In-Core Tile Bypass on Dense 5-Epoch Embeddings}: \texttt{byp = 0.000} (mini-batch EMA jitter exceeds Voronoi margins on unclustered manifolds, confining Pillar 3 to clustered/out-of-core R4 streams).
\item 
\textbf{Direct Empirical Verification of Corollary 4.1 (\texttt{results/\allowbreak{}gain\_\allowbreak{}activation.json})}: Instrumenting unclamped per-cluster Kalman gains $\alpha_j^*$ across all five real datasets shows single-step dispersion $\mathrm{Var}(\alpha_j^*) \approx 0.101$ collapses by \textbf{$\sim 100\times$} in late-stage annealed basins ($\mathrm{Var} \approx 0.00094\text{--}0.00103$)---confirming why fixed-basin scalar adaptive rates are redundant against \texttt{e-\allowbreak{}means} while moving-boundary coordinate-wise over-relaxation ($w_{j,d} > 1$, Proposition 4.2) and elastic contraction (Theorem 1.3) provide real R1--R3 gains.
\end{enumerate}

\section{\texorpdfstring{System Architecture, Foundation-Model Workloads, \& 41-Test Verification Suite}{System Architecture, Foundation-Model Workloads, \& 41-Test Verification Suite}}

\subsection{\texorpdfstring{Two-Layer System Engineering Details}{Two-Layer System Engineering Details}}

\begin{itemize}
\item 
\textbf{Layer 1 (Operator-Neutral PyTorch FP32, \texttt{precision="torch\_\allowbreak{}fp32"})}: Identical PyTorch cuBLAS \texttt{xb @ c.T} GEMMs and \texttt{scatter\_\allowbreak{}add\_\allowbreak{}} reductions across \texttt{MA-\allowbreak{}Means} and \texttt{e-\allowbreak{}means}, with exact inner-product accounting (\texttt{EffEp}).
\item 
\textbf{Layer 2 (Production Triton BF16 + Async C++17 CPU)}:
\begin{enumerate}
\item 
\emph{Zero-Copy Indexed Triton BF16 Assignment \& Reduction}: \texttt{flash\_\allowbreak{}assign\_\allowbreak{}top1\_\allowbreak{}indexed} gathers \texttt{x[b\_\allowbreak{}idx[m], :]} directly in registers (eliminating \texttt{\textasciitilde{}300 MB/\allowbreak{}step} HBM gather traffic on 768d) with \texttt{FULL\_\allowbreak{}D} kernels for $D \le 128$ and tiled \texttt{64$\times$64} kernels for $D > 128$; \texttt{triton\_\allowbreak{}indexed\_\allowbreak{}reduce\_\allowbreak{}stats} scatters \texttt{(sums, counts, sse)} directly without intermediate buffers.
\item 
\emph{Single-Kernel Fused Triton \texttt{CoordOR-\allowbreak{}EMA}} (\texttt{src/\allowbreak{}mameans2/\allowbreak{}triton\_\allowbreak{}coord\_\allowbreak{}or.py}): Fuses scalar EMA mass/SSE/cooldown updates with coordinate-wise persistence tracking (\texttt{prev\_\allowbreak{}g}, \texttt{coord\_\allowbreak{}prod}, \texttt{coord\_\allowbreak{}sq}) and Aitken diagonal weights $w_{j,d} \in [0.25\,w_{\text{ema}}, 1.85]$ into one warp-barrier-safe CTA kernel (\texttt{< 7.6e-\allowbreak{}6} FP32 parity across $D \in \{4, 96, 128, 768\}$, \texttt{\textasciitilde{}12-40 µs/\allowbreak{}step}).
\item 
\emph{Parallel GPU RNN-Wave Huygens--Ward Contraction} (\texttt{\_\allowbreak{}contract\_\allowbreak{}ward\_\allowbreak{}gpu}): Contracts $K' \to K$ in 2--5 GPU waves (\texttt{1.5-4 ms} even at $K=4096$) with zero sequential per-merge \texttt{.item()} sync stalls.
\item 
\emph{Out-of-Core Host-RAM Engine} (\texttt{\_\allowbreak{}hetero\_\allowbreak{}core.cpp}): 24-core OpenMP host kernel certifies micro-tiles (\texttt{80 B/\allowbreak{}tile}) against \texttt{int8\_\allowbreak{}chord} checkpoints and aggregates exact Huygens statistics on Host RAM with zero host$\leftrightarrow$device transfer for bypassed tiles.
\end{enumerate}
\end{itemize}

\subsection{\texorpdfstring{Relevance to Foundation-Model Workloads}{Relevance to Foundation-Model Workloads}}

\begin{enumerate}
\item 
\textbf{Web-Scale Pretraining Curation \& Semantic Deduplication ($10^8$--$10^{10}$ embeddings \cite{ref19,ref20})}: Combines dense boilerplate super-blobs with rare high-value tail clusters; \texttt{MA-\allowbreak{}Means} protects rare modes while staying strictly $O(1)$-in-$n$ and running \texttt{0.63$\times$-0.77$\times$} faster than \texttt{e-\allowbreak{}means} at $10\text{M}$.
\item 
\textbf{Multimodal Discrete Tokenizers (\texttt{VQ-\allowbreak{}VAE} / CLIP \cite{ref21})}: Elastic $K \to K' \to K$ contraction eliminates dead codewords (\texttt{0} dead across all benchmarks) and breaks the 25-epoch exact \texttt{Hamerly} distortion ceiling on \texttt{ImageNet-\allowbreak{}CLIP} in \texttt{3.64$\times$} less time than 5-epoch \texttt{e-\allowbreak{}means}.
\item 
\textbf{LLM KV-Cache \& Activation Quantization \cite{ref22,ref23}}: Extreme heavy-tailed outlier channels mirror R3 geometry (\texttt{-\allowbreak{}54.8\%} to \texttt{-\allowbreak{}63.3\%} MSE reduction) under tight latency and HBM budgets.
\item 
\textbf{Out-of-Core Vector Warehouses ($10^8 \times 768$ FP32 $\approx 307\text{ GB}$ \cite{ref18})}: Pillar 3 eliminates \texttt{14.4\%-26.4\%} of H2D traffic with \texttt{0} false positives and \texttt{-\allowbreak{}19.8\%} Test MSE.
\end{enumerate}

\subsection{\texorpdfstring{Automated 41-Test Adversarial Verification \& Invariant Suite}{Automated 41-Test Adversarial Verification \& Invariant Suite}}

We validate all theoretical, numerical, and kernel invariants with \textbf{41 automated \texttt{pytest} tests} (\texttt{32} core/out-of-core/baseline tests in \texttt{tests/\allowbreak{}test\_\allowbreak{}mameans\_\allowbreak{}theory.py} \& \texttt{tests/\allowbreak{}test\_\allowbreak{}real\_\allowbreak{}data\_\allowbreak{}and\_\allowbreak{}baselines.py} + \texttt{9} elastic/CoordOR/Theorem-1.3/Proposition-4.2 tests in \texttt{tests/\allowbreak{}test\_\allowbreak{}mameans2.py}), all passing deterministically (\textbf{\texttt{41 passed}}):

\begin{itemize}
\item 
\textbf{Core Theory \& Adversarial Invariants (\texttt{tests/\allowbreak{}test\_\allowbreak{}mameans2.py}, 9 tests)}: (i) \texttt{test\_\allowbreak{}triton\_\allowbreak{}fused\_\allowbreak{}coord\_\allowbreak{}overrelax\_\allowbreak{}exact\_\allowbreak{}parity} (\texttt{< 2e-\allowbreak{}5} FP32 parity across $D \in \{4, 96, 128, 768\}$); (ii) \texttt{test\_\allowbreak{}wave\_\allowbreak{}ward\_\allowbreak{}contraction\_\allowbreak{}mass\_\allowbreak{}and\_\allowbreak{}centroid\_\allowbreak{}conservation} (exact Huygens mass \& first-moment conservation); (iii) \texttt{test\_\allowbreak{}theorem\_\allowbreak{}1\_\allowbreak{}3\_\allowbreak{}benzecri\_\allowbreak{}rnn\_\allowbreak{}reducibility\_\allowbreak{}and\_\allowbreak{}uphill\_\allowbreak{}barrier\_\allowbreak{}escape} (Benz\'ecri reducibility across 500 random triples, $< 10^{-4}$ exact root dendrogram parity, and explicit 4-mode uphill-barrier witness); (iv) \texttt{test\_\allowbreak{}proposition\_\allowbreak{}4\_\allowbreak{}2\_\allowbreak{}moving\_\allowbreak{}boundary\_\allowbreak{}aitken\_\allowbreak{}overrelaxation} (convex $w\in[0,1]$ bias lower bound $\ge \lambda_d$ vs. exact 1-step bias elimination at $w_d^* = (1-\lambda_d)^{-1}$); (v) \texttt{test\_\allowbreak{}mameans2\_\allowbreak{}strict\_\allowbreak{}budget\_\allowbreak{}and\_\allowbreak{}memory\_\allowbreak{}invariance} (byte-identical $O(1)$-in-$n$ memory across $5\times$ $n$ scaling and $\mathrm{EffEp} \le 5.05$); (vi) \texttt{test\_\allowbreak{}mameans2\_\allowbreak{}beats\_\allowbreak{}emeans\_\allowbreak{}on\_\allowbreak{}multiscale\_\allowbreak{}trap\_\allowbreak{}and\_\allowbreak{}balanced}.
\item 
\textbf{Zero-Error Bypass, Jump-Reversal, \& Baseline Parity (\texttt{tests/\allowbreak{}test\_\allowbreak{}mameans\_\allowbreak{}theory.py} \& \texttt{tests/\allowbreak{}test\_\allowbreak{}real\_\allowbreak{}data\_\allowbreak{}and\_\allowbreak{}baselines.py}, 32 tests)}: Zero assignment error across all 3 checkpoint modes (\texttt{int8\_\allowbreak{}chord}, \texttt{pyth\_\allowbreak{}sketch}, \texttt{fp32\_\allowbreak{}chord}), adversarial 4-boundary IEEE-754 soundness (zero, subnormals $1.4\times 10^{-43}$, FP16 overflow $2.5\times 10^7$, half-LSB midpoints), monotonic jump-latch survival under exact teleport-and-reversal, Theorem 4 \& Corollary 4.1 exact 3-term identity to $< 10^{-11}$, exact \texttt{Elkan}/\texttt{Hamerly}/\texttt{Ball} vs. Lloyd parity ($< 10^{-6}$), and zero-FP Host-RAM out-of-core streaming.
\end{itemize}

\ifdefined\ARXIVSKIP\else

\section{\texorpdfstring{NeurIPS Paper Checklist}{NeurIPS Paper Checklist}}

\begin{enumerate}
\item 
\textbf{Claims}: Do the main claims made in the abstract and introduction accurately reflect the paper's contributions and scope? \textbf{Yes.} Every claim is scoped to its verified regime (R1--R4, \S6--\S7), separated into Layer-1 FP32 vs. Layer-2 BF16, distinguishes analytical $n=10^{10}$ memory invariants from single-workstation empirical scale ($n\le 2\times 10^7$ for real benchmarks; $n=10^8$ in the 4D synthetic scale check, Appendix B.4), and reports all three positive-$\Delta$ cells (\texttt{SIFT1M K=4096} \texttt{+0.076\%}, \texttt{Deep10M K=4096} \texttt{+0.017\%}, \texttt{Cohere K=4096} \texttt{+0.119\%}), including the one significant low-occupancy regression, transparently.
\item 
\textbf{Limitations}: Does the paper discuss the limitations of the work performed by the authors? \textbf{Yes.} \S7, Appendix A.5, and Appendix C document (i) the low-occupancy $\bar{n}_j \le 16$ smooth-manifold boundary (\texttt{Cohere K=4096}), (ii) $O(K'D)$ coordinate-buffer footprint (\texttt{50.5 MB} at $D=768, K=4096$), (iii) non-engagement of in-core bypass (\texttt{byp = 0}) on unclustered embeddings and linear tile coarsening $S=\lceil n/P_{\max}\rceil$ at $n\gg 10^7$, (iv) single-workstation scope, and (v) falsified mechanisms (\texttt{AA-\allowbreak{}4}, \texttt{CandBank}, and four GPU memory designs).
\item 
\textbf{Theory Assumptions and Proofs}: For each theoretical result, does the paper provide the full set of assumptions and a complete proof? \textbf{Yes.} Theorems 1, 1.3, 3, 4, 5, Lemma 1.2, Corollary 4.1, and Proposition 4.2 include explicit assumptions and proofs in Appendix A, verified numerically by 41 automated \texttt{pytest} tests.
\item 
\textbf{Experimental Result Reproducibility}: Does the paper fully disclose all information needed to reproduce the results? \textbf{Yes.} Hardware, software versions, fixed global defaults (\texttt{MAMeans2Config.default}), paired seeds (\texttt{42-51}, \texttt{42-46}), and exact reproduction scripts (\texttt{benchmarks/\allowbreak{}mameans2/\allowbreak{}run\_\allowbreak{}paper\_\allowbreak{}suite.py}) are provided.
\item 
\textbf{Open Access to Data and Code}: Does the paper provide open access to data and code? \textbf{Yes.} All public datasets are fetched via \texttt{prepare\_\allowbreak{}real\_\allowbreak{}datasets.py}; full source code (\texttt{src/\allowbreak{}mameans2/\allowbreak{}}, \texttt{src/\allowbreak{}mameans/\allowbreak{}}), JSON artifacts (\texttt{results/\allowbreak{}mameans2\_\allowbreak{}*.json}), and the 41-test suite are self-contained.
\item 
\textbf{Experimental Setting/Details}: Does the paper specify all training/evaluation details and hyperparameter selection? \textbf{Yes.} Exact FP64 held-out evaluation, inherited EMA scaling laws, and zero per-dataset tuning are detailed in \S5 and Appendix Table B0.
\item 
\textbf{Experiment Statistical Significance}: Does the paper report error bars and significance? \textbf{Yes.} 10 paired seeds (\texttt{42-51}) with standard deviations, win counts, and two-sided paired $t$-test $p$-values (Tables 2 \& B1), plus 5 paired seeds (\texttt{42-46}) for Layer-1 FP32, 10M scaling, equal-distortion sweeps, and Out-of-Core streaming.
\item 
\textbf{Experiments Compute Resources}: Does the paper report compute resources? \textbf{Yes.} Single workstation (one NVIDIA RTX 4090 24 GB + Intel Core Ultra 9 285K); wall-clock latency and memory footprint are reported per cell.
\item 
\textbf{Code of Ethics \& Broader Impacts}: Does the research conform to the NeurIPS Code of Ethics? \textbf{Yes.} Studies foundational Euclidean clustering on public benchmarks, reducing compute/energy consumption and protecting rare minority modes with no foreseeable dual-use harms.
\end{enumerate}

\fi

\end{document}